\documentclass{article}

\usepackage[preprint]{corl_2023} 
\usepackage[usenames,dvipsnames]{xcolor}
\definecolor{deepblue}{rgb}{0,0,0.5}
\definecolor{deepred}{rgb}{0.6,0,0}
\definecolor{deepgreen}{rgb}{0,0.5,0}
\definecolor{gray}{rgb}{0.4,0.4,0.4}
\definecolor{lightgreen}{rgb}{0.8,0.93,0.8}

\usepackage{graphicx}
\usepackage{booktabs}
\usepackage{makecell}
\usepackage{amsmath}
\usepackage{siunitx}
\usepackage{caption}
\usepackage{enumitem}
\usepackage{amsfonts}

\usepackage[scaled=0.85]{FiraMono}
\usepackage[T1]{fontenc}

\usepackage{subcaption}
\usepackage[subtle]{savetrees}
\usepackage[export]{adjustbox}
\usepackage{hyperref}
\usepackage{xcolor}
\usepackage[breakable]{tcolorbox}
\usepackage{listings}

\usepackage[capitalise, nameinlink]{cleveref}
\usepackage{setspace}

\newcommand{\langtorew}{Reward Translator}
\newcommand{\rewtoact}{Motion Controller}

\title{Language to Rewards for Robotic Skill Synthesis }

\author{%
Wenhao~Yu\thanks{Co-first authors, equal contribution},~ Nimrod~Gileadi$^*$, Chuyuan~Fu\thanks{Core contributors},~ Sean~Kirmani$^\dagger$, Kuang-Huei Lee$^\dagger$, \\
\textbf{Montse~Gonzalez~Arenas, Hao-Tien~Lewis~Chiang, Tom~Erez, Leonard~Hasenclever, }\\
\textbf{Jan~Humplik, Brian~Ichter, Ted~Xiao, Peng~Xu, Andy~Zeng, Tingnan~Zhang,}\\
\textbf{Nicolas Heess, Dorsa~Sadigh, Jie~Tan, Yuval~Tassa, Fei~Xia} \\
Google DeepMind
\\
\href{https://language-to-reward.github.io/}{https://language-to-reward.github.io/} \thanks{Corresponding emails: \href{mailto:magicmelon@google.com,nimrod@google.com,xiafei@google.com}{\texttt{\{magicmelon,nimrod,xiafei\}@google.com}} \newline See Contributions in Appendix \ref{app:author_contributions}}
\vspace{-1.0em}
}

\begin{document}
\maketitle

\hypersetup{
    linkcolor=MidnightBlue,
    filecolor=OliveGreen,
    urlcolor=MidnightBlue,
    citecolor=MidnightBlue,
    }


\setstretch{0.97}

\begin{abstract}
Large language models (LLMs) have demonstrated exciting progress in acquiring diverse new capabilities through in-context learning, ranging from logical reasoning to code-writing. Robotics researchers have also explored using LLMs to advance the capabilities of robotic control. However, since low-level robot actions are hardware-dependent and underrepresented in LLM training corpora, existing efforts in applying LLMs to robotics have largely treated LLMs as semantic planners or relied on human-engineered control primitives to interface with the robot. On the other hand, reward functions are shown to be flexible representations that can be optimized for control policies to achieve diverse tasks, while their semantic richness makes them suitable to be specified by LLMs.
In this work, we introduce a new paradigm that harnesses this realization by utilizing LLMs to define reward parameters that can be optimized and accomplish variety of robotic tasks. Using reward as the intermediate interface generated by LLMs, we can effectively bridge the gap between high-level language instructions or corrections to low-level robot actions. Meanwhile, combining this with a real-time optimizer, MuJoCo MPC, empowers an interactive behavior creation experience where users can immediately observe the results and provide feedback to the system.
To systematically evaluate the performance of our proposed method, we designed a total of 17 tasks for a simulated quadruped robot and a dexterous manipulator robot. We demonstrate that our proposed method reliably tackles $90\%$ of the designed tasks, while a baseline using primitive skills as the interface with Code-as-policies achieves $50\%$ of the tasks.
We further validated our method on a real robot arm where complex manipulation skills such as non-prehensile pushing emerge through our interactive system.

\end{abstract}


\section{Introduction}

The recent advancements in large language models (LLMs) pretrained on extensive internet data \cite{chowdhery2022palm, brown2020language} has revolutionized the ability to interpret and act on user inputs in natural language. These LLMs exhibit remarkable adaptability to new contexts (such as APIs \cite{liang2022code}, task descriptions \cite{zeng2022socratic}, or textual feedback \cite{huang2022inner}), allowing for tasks ranging from logical reasoning \cite{wei2022chain, kojima2022large} to code generation \cite{chen2021evaluating} with minimal hand-crafted examples.

\begin{figure}[t]
    \centering
    \includegraphics[width=0.98\textwidth]{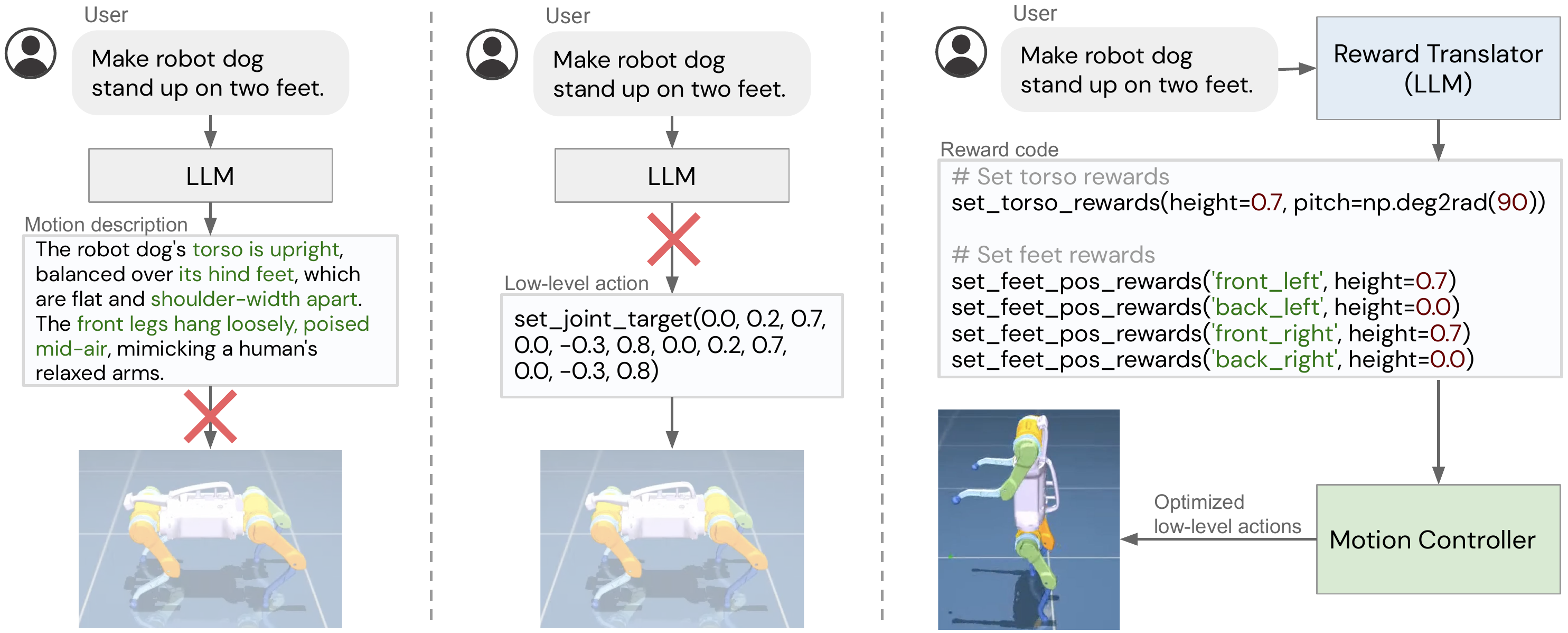}
    \caption{LLMs have some internal knowledge about robot motions, but cannot directly translate them into actions (left). Low-level action code can be executed on robots, but LLMs know little about them (mid). We attempt to bridge this gap, by proposing a system (right) consisting of the \langtorew that interprets the user input and transform it into a reward specification. The reward specification is then consumed by a Motion Controller that interactively synthesizes a robot motion which optimizes the given reward.}
    \label{fig:l2r_overview}
    \vspace{-0.15in}
\end{figure}

These diverse applications have extended to the field of robotics as well, where substantial progress has been made in using LLMs to drive robot behaviors \cite{ liang2022code, huang2022inner, zeng2022socratic, ahn2022can, vemprala2023chatgpt, snell2022context}: from step-by-step planning \cite{zeng2022socratic, ahn2022can, huang2022language}, goal-oriented dialogue \cite{vemprala2023chatgpt,snell2022context}, to robot-code-writing agents \cite{liang2022code, singh2022progprompt}. While these methods impart new modes of compositional generalization, they focus on using language to concatenate together new behaviors from an existing library of control primitives that are either manually-engineered or learned a priori.
Despite having internal knowledge about robot motions, LLMs struggle with directly outputting low-level robot commands due to the limited availability of relevant training data (\cref{fig:l2r_overview}).
As a result, the expression of these methods are bottlenecked by the breadth of the available primitives, the design of which often requires extensive expert knowledge
or massive data collection~\cite{jang2021bc,brohan2022rt,lynch2022interactive}.

To tackle these challenges, we need to operate at a level of abstraction that allows harnessing the intuitive and interactive capabilities offered by LLMs. 
Our key insight is to leverage reward functions as an interface that bridges the gap between language and low-level robot actions.
This is motivated by the fact that language instructions from humans often tend to describe behavioral outcomes instead of low-level behavioral details (e.g. ``robot standing up'' versus ``applying 15 Nm to hip motor''), and therefore we posit that it would be easier to connect instructions to rewards than low-level actions given the richness of semantics in rewards.
In addition, reward terms are usually modular and compositional, which enables concise representations of complex behaviors, goals, and constraints. This modularity further creates an opportunity for the user to interactively steer the robot behavior.
However, in many previous works in reinforcement learning (RL) or model predictive control (MPC), manual reward design requires extensive domain expertise \cite{lee2019robust, siekmann2021sim, xia2020relmogen}. 
While reward design can be automated, these techniques are  sample-inefficient and still requires manual specification of an objective indicator function for each task \cite{chiang2019learning}. 
This points to a missing link between the reward structures and task specification which is often in natural language. As such, we propose to utilize LLMs to automatically generate rewards, and leverage online optimization techniques to solve them. Concretely, we explore the code-writing capabilities of LLMs to translate task semantics to reward functions, and use MuJoCo MPC, a real-time optimization tool to synthesize robot behavior in real-time~\cite{howell2022}. 
Thus reward functions generated by LLMs can enable non-technical users to generate and steer novel and intricate robot behaviors without the need for vast amounts of data nor the expertise to engineer low-level primitives.

The idea of grounding language to reward has been explored by prior work for extracting user preferences and task knowledge  ~\cite{goyal2019using,lin2022inferring,fan2022minedojo,sharma2022correcting,kwon2023reward,hu2023language}. Despite promising results, they usually require training data to learn the mapping between language to reward. With our proposed method, we enable a data efficient interactive system where the human engages in a dialogue with the LLM to guide the generation of rewards and, consequently, robot behaviors (\cref{fig:l2r_overview}).

Across a span of 17 control problems on a simulated quadruped and a dexterous manipulator robot, we show that this formulation delivers diverse and challenging locomotion and manipulation skills. Examples include getting a quadruped robot to stand up, asking it to do a moonwalk, or tasking a manipulator with dexterous hand to open a faucet. We perform a large-scale evaluation to measure the overall performance of our proposed method. We compare our method to a baseline that uses a fixed set of primitive skills and an alternative formulation of grounding language to reward. We show that our proposed formulation can solve $40\%$ more skills than baselines and is more stable in solving individual skills.
We further deploy our approach to a real robot manipulator and demonstrate complex manipulation skills through language instructions.


\section{Related Work}
\label{sec:related-work}
Here we discuss relevant prior work that reason about language instructions to generate robot actions, code, or rewards to ground natural language to low-level robotic control. We then discuss work focused on responding to interactive human feedback such as language corrections.

\noindent\textbf{Language to Actions.}
Directly predicting low-level control actions based on a language instruction has been studied using various robot learning frameworks.
Early work in the language community studied mapping templated language to controllers with temporal logic~\cite{kress2008translating} or learning a parser to motion primitives~\cite{matuszek2013learning}, while more recent work utilize end-to-end models that produce actions conditioned on natural language descriptions. One example is instruction following methods in navigation~\cite{ku2020room}. However, they often assume low-dimensional actions navigating from one node of the graph to another~\cite{ku2020room, kamath2023new}.
To extend the end-to-end approaches to manipulation, a common approach is to utilize latent embeddings of language commands as multitask input context, and train with behavioral cloning \cite{jang2021bc,mees23hulc2,lynch2022interactive}, offline reinforcement learning \cite{ebert2021bridge}, goal-conditioned reinforcement learning \cite{fu2019language}, or in a shared autonomy paradigm~\cite{karamcheti2021lila}. While end-to-end trained policies can be performant, they require significant amount of data in the form of offline datasets or online environment interaction. In contrast, we study a less data hungry approach where low-level actions are not directly produced by an end-to-end policy but instead by an optimal controller.

\noindent \textbf{Language to Code.}
Code generation models have been widely studied both in and outside robotics context~\cite{austin2021program,chen2021evaluating,ellis2020dreamcoder}. The capability of those models range from solving coding competition questions~\cite{li2022competition} and benchmarks~\cite{alet2021large}, to drawing simple figures~\cite{tian2020learning}, generating policies that solve 2D tasks~\cite{trivedi2021learning}, and complex instruction following tasks~\cite{liang2022code}.
In this work, we study LLMs for generating code for reward functions, and show that the expression of the rewards can lead to expressive low-level policies.

\noindent \textbf{Language to Rewards.}
The idea of translating natural language instructions to rewards has been explored by several prior work \cite{kwon2023reward, lin2022inferring, sharma2022correcting, pmlr-v164-nair22a, goyal2019using, bahdanau2018learning, hu2023language}. A common strategy in this direction is to train domain-specific reward models that map language instructions to reward values \cite{lin2022inferring, goyal2019using, pmlr-v164-nair22a} or constraints \cite{sharma2022correcting}. Although these methods can achieve challenging language conditioned robotic tasks such as object pushing \cite{sharma2022correcting}, and drawer opening \cite{pmlr-v164-nair22a}, they require considerable language-labeled data to train the reward model. Recent works investigated using LLMs directly as a reward function for inferring user intentions in negotiation games or collaborative human-AI interaction games \cite{kwon2023reward, hu2023language}. By leveraging LLMs to assign reward values during RL training, they demonstrate training agents that are aligned with user intentions and preferences without explicit reward modeling. However, these works receive reward values of rollouts when training RL policies, which requires a large number of queries to LLMs during training. In contrast, we levrage LLMs to produce a parameterized reward function that can then be optimized. A similar direction to this work is automated parameterization of reward functions, which had been explored in AutoRL \cite{chiang2019learning}, however, they don't provide a language interface.

\noindent \textbf{Incorporating Iterative Human Feedback.}
Correcting plans with iterative language feedback has also been explored in the past.
Broad et al. enable efficient online corrections using distributed correspondence graphs to ground language~\cite{broad2017realtime}. However, this work relies on a semantic parser with pre-defined mappings to ground language corrections. More end-to-end approaches have also demonstrated learning a language correction conditioned policy, but they are similarly data hungry and thus fall back to shared autonomy to reduce complexity~\cite{cui2023no}.
Later work explore mapping language corrections to composable cost functions similar to our work by training a prediction model from demonstration and apply trajectory optimization to perform control~\cite{sharma2022correcting}. Followup works further simplifies the system by integrating language corrections to directly modify the waypoints of a trajectory using extensive datasets of paired corrections and demonstrations \cite{bucker2022reshaping,bucker2022latte}.
In contrast to these prior work, we demonstrate a flexible and data-efficient approach that leverages LLMs to allow for multi-step correction of reward functions based on human feedback.
\section{Grounding Language to Actions Using Rewards}
\label{sec:method}

\subsection{Background and Reward Interface}
Our system takes user instruction in natural language and synthesizes corresponding robot motions by leveraging reward function as the interface to communicate with low-level controllers. 
We define the reward function in the context of Markov Decision Process (MDP), commonly used to formulate robot control problems: $(S, A, R, P, p_0)$, where $S$ is the state space, $A$ is the action space, $R: S \times A \mapsto \mathbb{R}$ is the reward function, $P: S \times A \mapsto S$ is the dynamics equation, and $p_0$ is the initial state distribution. Given a reward function $R$, an optimal controller finds a sequence of actions $\mathbf{a}_{1:H} =\{ \mathbf{a}_1, \dots, \mathbf{a}_H\}$ that maximizes the expected accumulated reward: $J(\mathbf{a}_{1:H}) = \mathbb{E}_{\tau=(\mathbf{s}_0, \mathbf{a}_0, \dots, \mathbf{s}_H)} \sum_{t=0}^{H} R(\mathbf{s}_t, \mathbf{a}_t)$, where $H$ is the rollout horizon.

In this work, we assume the reward takes a particular form, suitable for use with MJPC (see below). The reward is the sum of a set of individual terms:
\setstretch{0.5}
\begin{equation}
{{R}}({\mathbf{s}}, {\mathbf{a}}) = -\sum_{i = 0}^M w_i \cdot \text{n}_i\big({{r}}_i({\mathbf{s}}, {\mathbf{a}}, \psi_i)\big),
\label{eq:mjpc_objective}
\end{equation}
\setstretch{1.0}
where $w \in \mathbf{R}_{+}$ is a non-negative weight, $\text{n}(\cdot) : \mathbf{R} \rightarrow \mathbf{R}_+$ is a twice-differentiable norm that takes its minimum at $0$, $r \in \mathbf{R}$ is a residual term that achieves optimality when $r=0$, and $\psi_i$ is the parameters of the $i_{th}$ residual term. For example, if we want to have the robot raise its body height $h$ to a desired height, we may design a residual term $r_h(h, \psi) = h-\psi$, where the reward parameter $\psi$ denotes the desired height, and use the l2 norm to construct the final reward function: $R_h = -w ||r_h||_2$. In principle, one may design task-specific residual terms that can solve particular controller tasks. However, designing these residuals requires domain expertise and may not generalize to novel tasks. In this work, we use a set of generic and simple residual terms, and leverage the power of LLMs to compose different terms to generate complex behaviors. 
The full set of residual terms used in this work can be found in the Appendix \ref{app:reward_terms}.

Our proposed system consists of two key components (\cref{fig:l2r_overview} right): i) a \emph{\langtorew}, built upon pre-trained Large Language Models (LLMs) \cite{vemprala2023chatgpt}, that interacts with and understands user intents and modulates all reward parameters $\psi$ and weights $w$, and ii) a \emph{\rewtoact}, based on MuJoCo MPC \cite{howell2022}, that takes the generated reward and interactively optimize the optimal action sequence $\mathbf{a}_{1:H}$. Below we provide more details on the design of \langtorew and \rewtoact.

\subsection{\langtorew}

\begin{figure}[h]
    \centering
    \includegraphics[width=0.99\textwidth]{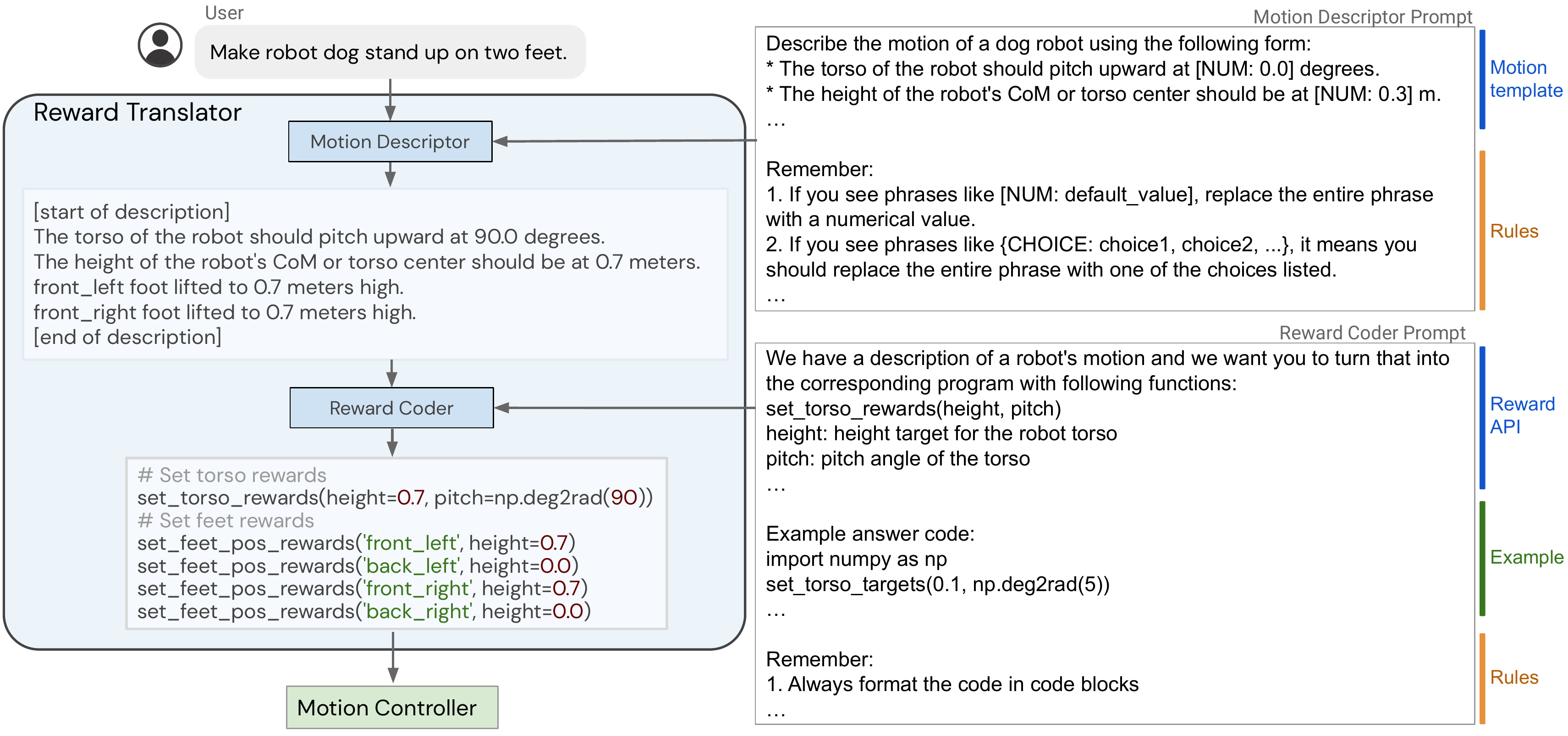}
    \caption{Detailed dataflow of the \langtorew. A \emph{Motion Descriptor} LLM takes the user input and describe the user-specified motion in natural language, and a \emph{Reward Coder} translates the motion into the reward parameters.}
    \label{fig:smd_workflow}
\end{figure}

Inspired by recent progress on Large Language Models (LLMs), we propose to build the \langtorew based on LLMs to map user interactions to reward functions corresponding to the desired robot motion. As reward tuning is highly domain-specific and requires expert knowledge, it is unsurprising that LLMs trained on generic language datasets (e.g. \cite{chowdhery2022palm}) cannot directly generate a reward for a specific hardware. 
Instead, we explore the in-context learning ability of LLMs to achieve this goal, inspired by prior work that demonstrated a variety of in-context learning skills for LLMs \cite{brown2020language, ziems2023large}. Furthermore, we decompose the problem of language to reward into two stages: motion description and reward coding task, as illustrated in \cref{fig:smd_workflow}.

\textbf{Motion Description }
In the first stage, we design a \emph{Motion Descriptor} LLM that interprets and expands the user input into a natural language description of the desired robot motion following a pre-defined template (see example in \cref{fig:smd_workflow}). Although it is possible for LLMs to directly generate reasonable reward functions for relatively simple task, it often fails for tasks that necessitates complex reasoning. On the other hand, as observed in \cref{fig:l2r_overview} left, LLMs can describe complex motions in detailed natural language successfully.

Inspired by this observation, we design a template that describes common movements of a robot (see~\cref{fig:smd_workflow} top right for an example of the template and the prompt for the LLM) to effectively harness LLMs' internal knowledge about motions. The role of the \emph{Motion Descriptor} is to complete the provided template (e.g., replacing certain elements such as \texttt{CHOICE} and \texttt{NUM} in the example.
This helps the \emph{Motion Descriptor} produce more structured and predictable outputs and improves stability of the overall system. In addition, as we are describing the motion in natural language, we do not need to provide any specific examples in the prompt and can rely entirely on LLMs to generate the result.

\textbf{Reward Coding } 
In the second stage, we translate the generated motion description into the reward function using a second LLM. We formulate the problem of language to reward function as a code-writing task to benefit from the LLMs' knowledge of coding and code structure, thus we name the second LLM the \emph{Reward Coder}. We design a prompt for instructing the LLM to generate reward specifying code (see example in \cref{fig:smd_workflow} bottom right). The prompt consists of three parts: i)  description of the reward APIs that the LLM can call to specify different parameters of the reward function, ii) an example response that we expect the \emph{Reward Coder} to produce, and iii) the constraints and rules that the \emph{Reward Coder} needs to follow. Note that the example is to demonstrate to the LLM how the response should look like, instead of teaching it how to perform a specific task. As such, the \emph{Reward Coder} needs to specify the reward parameters based on its own knowledge about the motion from the natural language description.

\subsection{\rewtoact}

The \rewtoact~needs to map the reward function generated by the \langtorew to low-level robot actions $\mathbf{a}_{1:H}$ that maximize the accumulated reward $J(\mathbf{a}_{1:H})$. There are a few possible ways to achieve this, including using reinforcement learning (RL), offline trajectory optimization, or, as in this work, receding horizon trajectory optimization, i.e., model predictive control (MPC). At each control step, MPC plans a sequence of optimized actions $\mathbf{a}_{1:H}$ and sends to the robot. The robot applies the action corresponding to its current timestamp, advances to the next step, and sends the updated robot states to the MJPC planner to initiate the next planning cycle. The frequent re-planning in MPC empowers its robustness to uncertainties in the system and, importantly, enables interactive motion synthesis and correction. Specifically, we use an open-source implementation based on the MuJoCo simulator \cite{todorov2012mujoco}, MJPC \cite{howell2022}. MJPC has demonstrated the interactive creation of diverse behaviors such as legged locomotion, grasping, and finger-gaiting while supporting multiple planning algorithms, such as iLQG and Predictive Sampling. Following the observation by Howell et al \cite{howell2022}, second-order planners such as iLQG produces smoother and more accurate actions while zeroth-order planners such as Predictive Sampling is better at exploring non-smooth optimization landscape, we use iLQG for legged locomotion tasks, while use Predictive Sampling for manipulation tasks in this work.

\section{Experiments}
\label{sec:result}

We design experiments to answer the following questions:

1) Is our proposed method, by combining LLMs and MJPC, able to generate diverse and complex robot motions through natural language interface?

2) Does interfacing with the reward function result in a more expressive pipeline than interfacing directly with low-level or primitive actions and is Motion Descriptor necessary for achieving reliable performance?

3) Can our method be applied to real robot hardware?

\subsection{Experiment Setup}

\begin{figure}[ht]
    \centering
    \begin{subfigure}[b]{0.48\textwidth}
    \includegraphics[width=\textwidth]{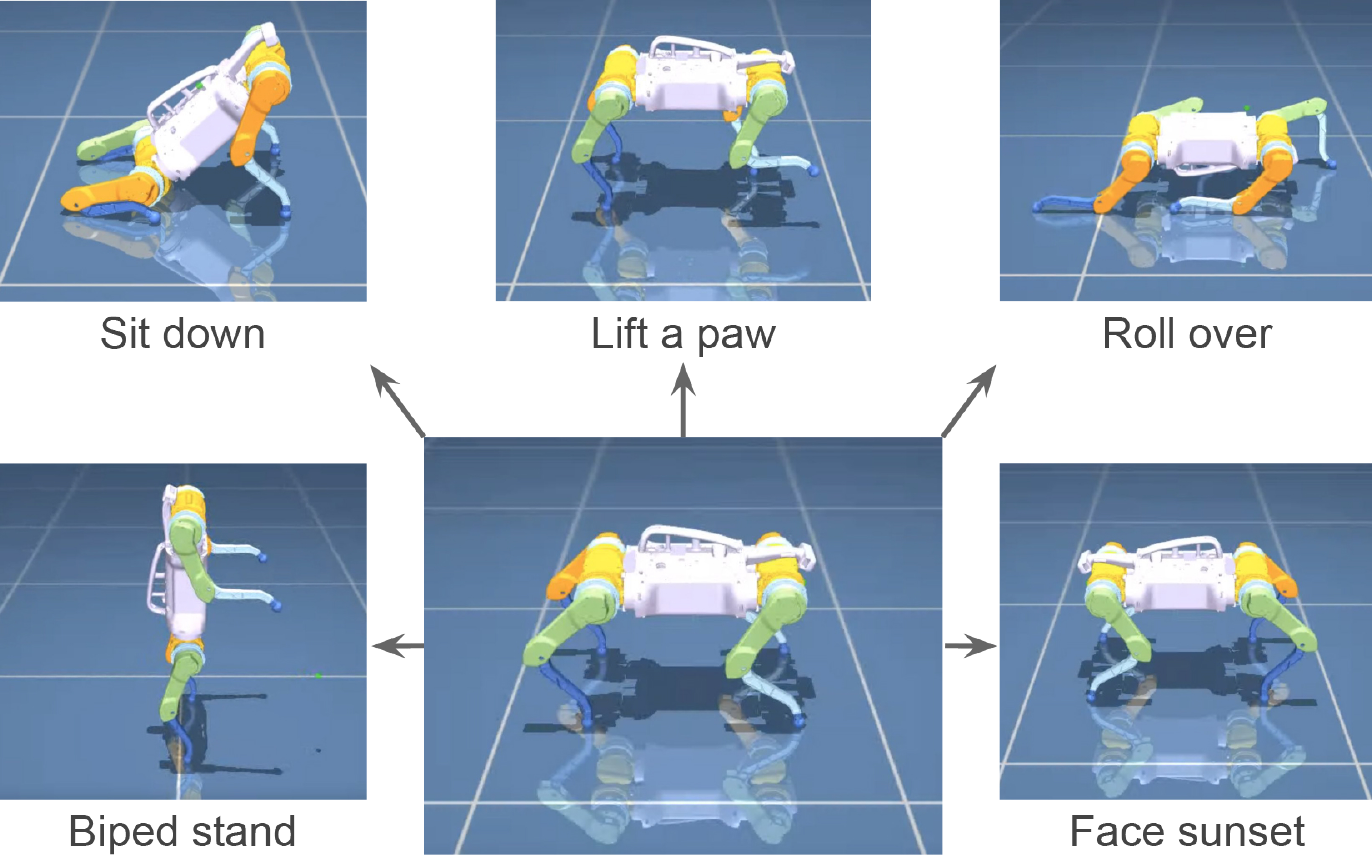}
    \caption{Quadruped robot}
    \label{fig:ap}
    \end{subfigure}%
    \hfill
    \begin{subfigure}[b]{0.48\textwidth}
    \includegraphics[width=\textwidth]{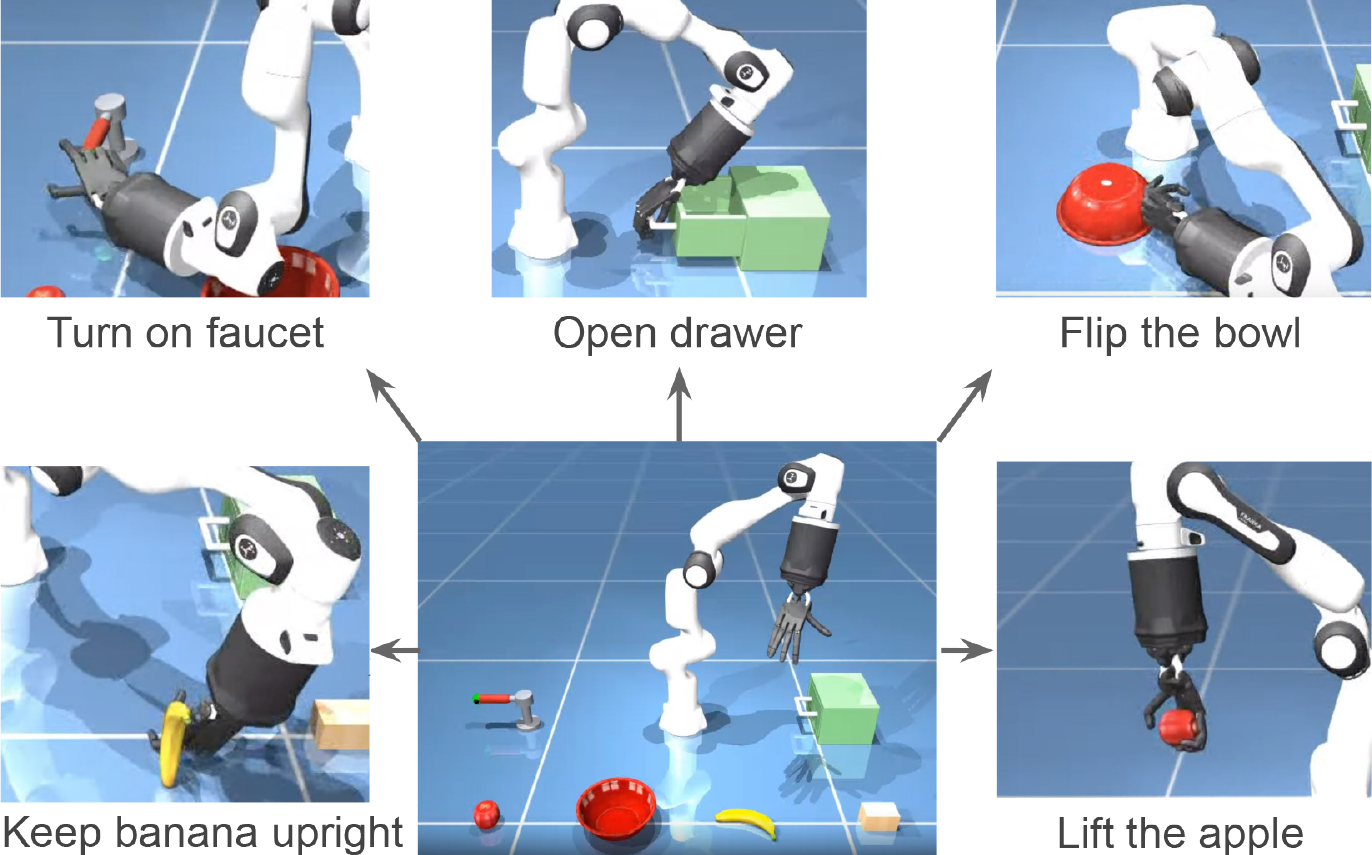}
    \caption{Dexterous manipulator robot}
    \label{fig:sf}
    \end{subfigure}%
    \\
    \begin{subfigure}[b]{\textwidth}
    \includegraphics[width=\textwidth]{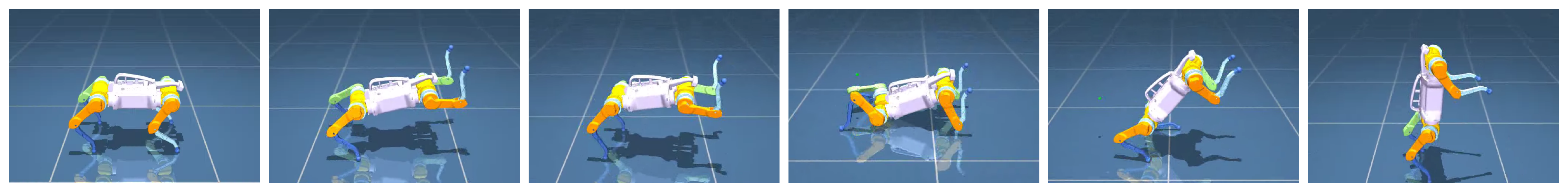}
    \end{subfigure}%
    \\
    \vspace{-0.15in}
    \begin{subfigure}[b]{\textwidth}
    \includegraphics[width=\textwidth]{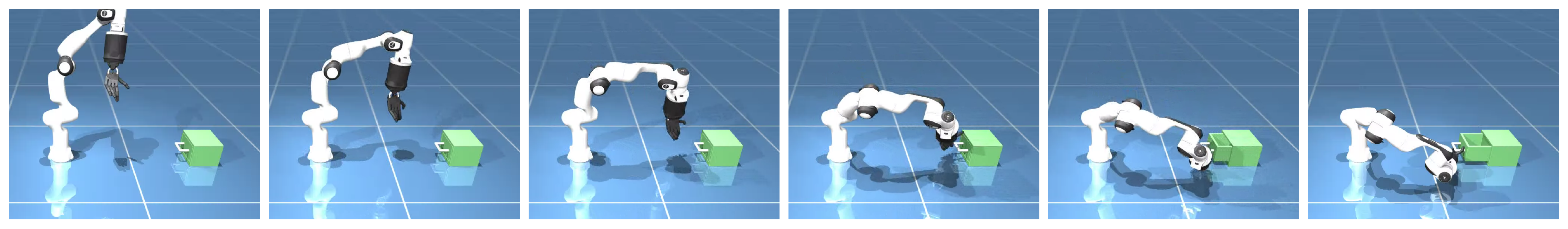}
    \caption{Example rollout for the two robots.}
    \end{subfigure}%
    \caption{The two robots used in our experiments and sampled tasks. (a) a Quadruped robot with 12 DoFs. (b) a dexterous manipulator robot with 27 DoFs. (c) example rollouts produced by our algorithm.}
    \label{fig:robots}
\end{figure}

We evaluate our approach on two simulated robotic systems: a quadruped robot, and a dexterous robot manipulator (\cref{fig:robots}). Both robots are modeled and simulated in MuJoCo MPC \cite{howell2022}. In all experiments we use GPT-4 as the underlying LLM module \cite{openai2023gpt}. Here we describe the key setups of each robot. More details regarding the full prompts and reward function can be found in Appendix \ref{app:full_prompts} and \ref{app:reward_terms}.

\textbf{Quadruped Robot }
In this example, we demonstrate using our system to command a four legged robot (\cref{fig:robots} (a)) to perform a variety of motor skills. The quadruped robot has $12$ joints, $3$ on each leg. Quadruped robots have been demonstrated to perform a large variety of skills including locomotion \cite{tian2020learning}, hopping \cite{siekmann2021sim}, biped standing \cite{smith2023learning, fuchioka2022opt}, parkour \cite{caluwaerts2023barkour}, etc. We apply our system to the quadruped robot to perform a similar suite of skills while only using natural language as input.

\textbf{Dexterous Manipulator }
In the second example, we demonstrate our system on a dexterous manipulator robot. The robot consists of a 7 DoF Franka Emika arm and a 20 DoF shadow hand as the end-effector (as shown in \cref{fig:robots} (b)). This creates a large action space, making it challenging to manually train a controller to directly perform tasks using this robot.

\subsection{Baselines}
\label{ssec:baseline}

We compare our proposed system to two baseline methods: i) an ablation of our approach that only uses \emph{Reward Coder} without having access to the \emph{Motion Descriptor}, and ii) Code-as-Policies  \cite{liang2022code} where the LLM generates a plan for the robot motion using a set of pre-defined robot primitive skills instead of reward functions. For the Code-as-Policies (CaP) baseline, we design the primitive skills based on common commands available to the robot. Due to limited space we put the full list of primitives in \cref{app:baseline_detail}.

\subsection{Tasks}

We design nine tasks for the quadruped robot and eight tasks for the dexterous manipulator to evaluate the performance of our system. \cref{fig:robots} shows samples of the tasks. The full list of tasks can be found in \cref{app:task lsit}. Videos of sampled tasks can also be found in supplementary video and project website \footnote{\url{language-to-reward.github.io}}.

For the quadruped robot, the tasks can be categorized into four types: \emph{1) Heading direction control}, where the system needs to interpret indirect instructions about the robot's heading direction and to control the robot to face the right direction (e.g., identify the direction of sunrise or sunset). \emph{2) Body pose control}, where we evaluate the ability of the system to understand and process commands to have the robot reach different body poses, inspired by common commands issued by human to dogs such as sit and roll over. \emph{3) Limb control}, where we task the system to lift particular foot of the robot. Furthermore, we also test the ability of the system to take additional instructions to modify an existing skill, such turn in place with lifted feet. \emph{4) Locomotion styles}, where we evaluate our proposed system in generating different locomotion styles. In particular, we design a challenging task of having the quadruped stand up on two back feet to walk in a bipedal mode.

For the dexterous manipulator, we design tasks to test its ability to achieve different types of interactions with objects such as lifting, moving, and re-orienting. We test the system on a diverse set of objects  with significantly different shapes and sizes (\cref{fig:robots}) for each task. We further include two tasks that involves interacting with articulated objects of different joint types.

    

\begin{figure}[t]
    \begin{subfigure}[t]{0.99\textwidth}
    \includegraphics[width=\textwidth]{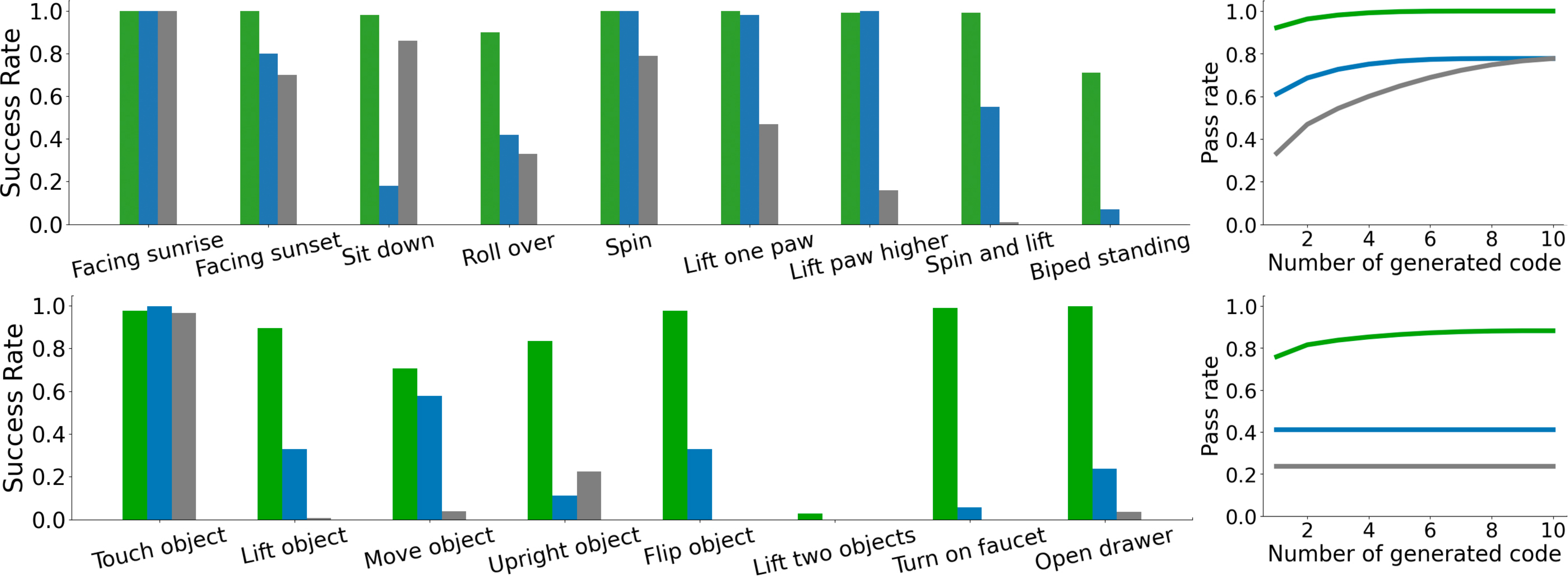}
    \end{subfigure}
    \\
    \includegraphics[width=0.5\textwidth, center]{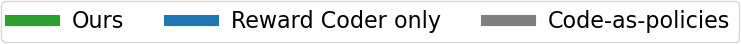}
    \caption{Comparison of our method and alternative methods in terms of pass rate: if we generate N pieces of code for each task and pick the best performing one, what's the percentage of tasks that the system can successfully tackle.}
    
    \label{fig:eval_result}
\end{figure}

\subsection{Evaluation results}
\label{ssec:eval}
For each task and method considered, we generate 10 responses from \langtorew, each evaluated in MJPC for $50$ times, thus we measure the end-to-end stability of the full pipeline. \cref{fig:eval_result} shows the results for both robots. Our proposed approach achieves significantly higher success rate for $11/17$ task categories and comparable performance for the rest tasks, showing the effectiveness and reliability of the proposed method. When compared to the CaP baseline, our method achieves better success rate in almost all tasks. This is due to that CaP can perform well on tasks that can be expressed by the given primitives (e.g. Touch object) or very close to the given examples in prompt (e.g. Sit down), but fails to generalize to novel low-level skills. On the other hand, using Reward Coder only can achieve success on some tasks but fails in ones that requires more reasoning. For example, when asked to open a drawer, the Reward Coder only baseline often forget to task the robot hand to get closer to the drawer handle and only design the reward for encouraging the drawer to be open. Sampled responses from different method can be found in \cref{app:example_outputs}.

To further understand the overall performance of different systems, we also show the pass rate in \cref{fig:eval_result} right, which is a standard metric for analyzing code generation performance \cite{chen2021evaluating}. For each point in the plot, it represents the percentage of tasks the system can solve, given that it can generate N pieces of code for each task and pick the best performing one. As such, the pass rate curve measures the stability of the system (the more flat it is, the more stable the system is) as well as the task coverage of the system (the converged point represents how many tasks the system can solve given sufficient number of trials). It is clear from the result that for both embodiments, using reward as the interface empowers LLMs to solve more tasks more reliably, and the use of Structured Motion Description further boosts the system performance significantly.


\subsection{Interactive Motion Synthesis Results}
\begin{figure}[h]
    \centering
    \includegraphics[width=\textwidth]{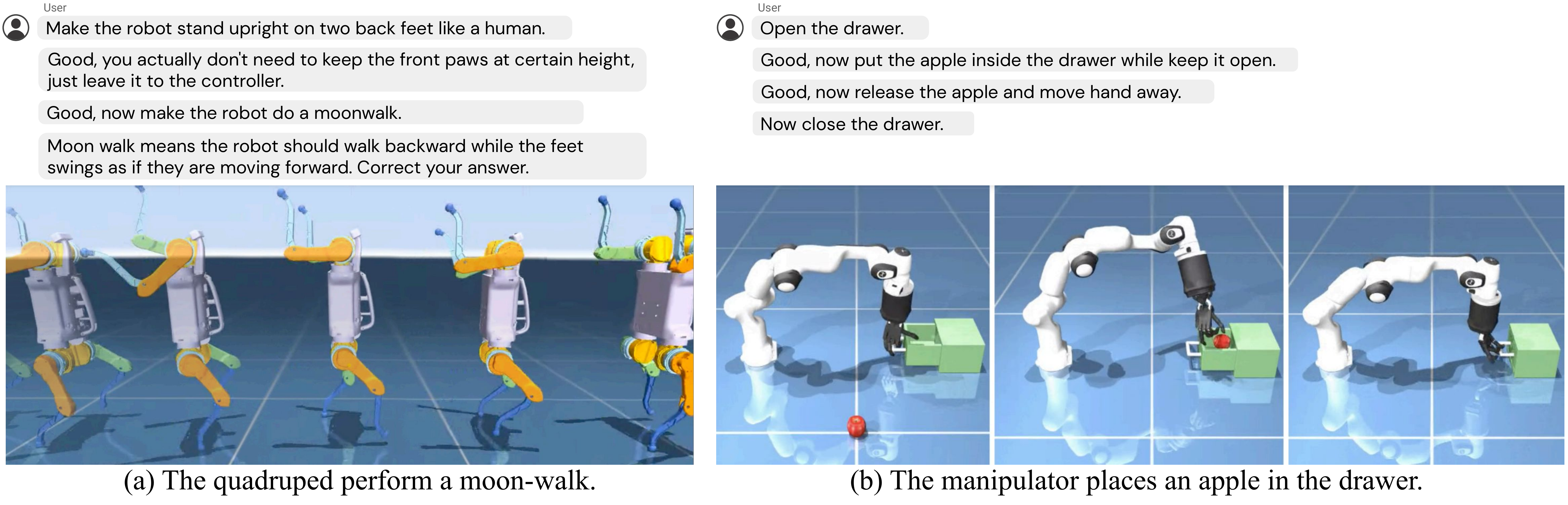}
    \caption{The two interactive examples using our proposed system.}
    \label{fig:interactive_robots}
\end{figure}

One benefit of using a real time optimization tool like MJPC is that humans can observe the motion being synthesized in real time and provide feedback. We showcase two examples where we teach the robot to perform complex tasks through multiple rounds of interactions. In the first example, we task the quadruped robot to stand up and perform a moon-walk skill (\cref{fig:interactive_robots}a). We give four instructions to achieve the task, as shown in \cref{fig:interactive_robots}. Each instruction improves the behavior towards the desired behavior based on the interactively synthesized results. This showcase that users can interactively shape the behavior of the robot in natural language.
In the second example, we showcase a different way of leveraging the interactivity of our system by sequentially commanding the dexterous manipulator robot to place an apple in a drawer, as seen in \cref{fig:interactive_robots}b.
Results of the interactive results are best viewed in the supplementary video and full code output from our method can be found in \cref{app:interact_code_output}.

\subsection{Real-robot experiments}
We implement a version of our method onto a mobile manipulator, and tested it on nonprehensile manipulation tasks in the real world.
\begin{figure}[h]
    \centering
    \includegraphics[width=\textwidth]{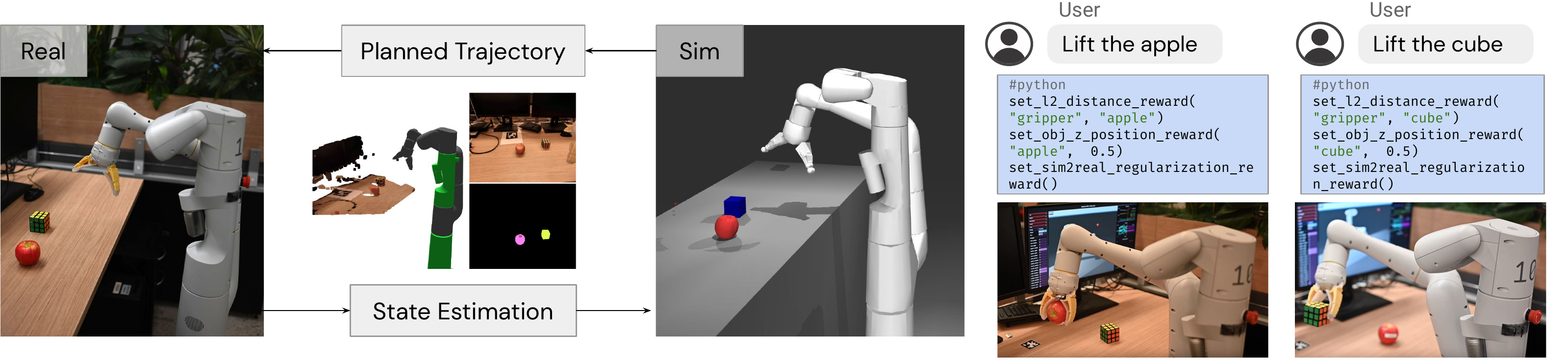}
    \caption{Implementation and rollouts of the proposed system in the real world.}
    \label{fig:l2r_real}
\end{figure}
In simulation, we have access to the ground-truth state for objects in the scene. In the real world, we detect objects in image-space using an open-vocabulary detector: F-VLM \cite{weicheng2023fvlm}. We extract the associated points from point cloud behind the mask and perform outlier rejection for points that might belong to the background. From a birds-eye view, we fit a minimum volume rectangle and take the extremes to determine the extent in the z-axis. We use this 3D bounding box as state estimation for corresponding object in simulation. To detect the surface of the table with proper orientation, we use an AprilTag \cite{olson2011april}. In addition, as seen in the supplementary video, MJPC can discover highly dexterous and dyanmic maneuvers to accomplish the desired task. However, these movements are beyond the capabilities of current real hardwares. To mitigate this issue, we design a regularization residual term specific to encourage steady and stable robot movements when applying our system to the real robot (\texttt{set$\_$sim2real$\_$regularization$\_$reward()} in \cref{fig:l2r_real}, see \cref{app:sim2real_term} for details for this term).

We demonstrate sim-to-real transfer on two tasks: object pushing and object grasping. Our system is able to generate relevant reward code and the Mujoco MPC is able to synthesize the pushing and grasping motion.  For rollouts please refer to the supplementary video/website and \cref{fig:l2r_real2}.
\section{Discussion and Conlusion}
\label{sec:conclusion}

In this work, we investigate a new paradigm for interfacing an LLM with a robot through reward functions, powered by a low-level model predictive control tool, MuJoCo MPC. Using reward function as the interface enables LLMs to work in a semantic-rich space that play to the strengths of LLMs, while ensures the expressiveness of the resulting controller. To further improve the performance of the system, we propose to use a motion description template to better extract internal knowledge about robot motions from LLMs. We evaluate our proposed system on two simulated robotic platforms: a quadruped robot and a dexterous manipulator robot. We apply our approach to both robots to acquire a wide variety of skills. Compared to alternative methods that do not use reward as the interface, or do not use the motion description template, our method achieves significantly better performance in terms of stability and the number of tasks it can solve.

\textbf{Limitations and Future Work.}
Though we show that our system can obtain a diverse set of skills through natural language interactions, there are a few limitations. 
First, we currently design templates of motion descriptions for each type of robot morphology, which requires manual work. An interesting future direction is to unify or automate the template design to make the system easily extendable to novel robot morphologies.
Second, our method currently relies on language as the interaction interface with human users. As such, it can be challenging to design tasks that are not easily described in language (e.g., ``walk gracefully"). One potential way to mitigate this issue is to extend the system to multi-modal inputs to allow richer forms of user interactions (e.g., by showing a video of the desirable behavior).
Thirdly, we currently use pre-defined reward terms whose weights and parameters are modulated by the LLMs. Constraining the reward design space helps improve stability of the system while sacrifices some flexibility. For example, our current design does not support time-varying rewards and would require re-designing the prompt to support that. Enabling LLMs to reliably design reward functions from scratch is thus an important and fruitful research direction.



\acknowledgments{The authors would like to acknowledge Ken Caluwaerts, Kristian Hartikainen, Steven Bohez, Carolina Parada, Marc Toussaint, and the greater teams at Google DeepMind for their feedback and contributions.}


\bibliography{main}  

\appendix
\clearpage

\section{Appendix}

\subsection{Author Contributions}
\label{app:author_contributions}

Author contributions by type, ordered alphabetically within each category:

\textbf{Method (conception, implementation, iteration, evaluation)}: Nimrod Gileadi, Kuang-Huei Lee, Yuval Tassa, Fei Xia, Peng Xu, Wenhao Yu.

\textbf{Infrastructure Development}: Tom Erez, Nimrod Gileadi, Yuval Tassa, Fei Xia, Wenhao Yu.

\textbf{Hardware Deployment}: Chuyuan Fu, Nimrod Gileadi, Leonard Hasenclever, Jan Humplik, Sean Kirmani, Yuval Tassa, Fei Xia, Ted Xiao, Wenhao Yu.

\textbf{Project Advising}: Tom Erez, Nicolas Heess, Brian Ichter, Dorsa Sadigh, Jie Tan, Yuval Tassa, Fei Xia, Andy Zeng, Tingnan Zhang. 

\textbf{Paper Writing/Revision}: Montse Gonzalez Arenas, Hao-Tien Lewis Chiang, Nimrod Gileadi, Brian Ichter, Dorsa Sadigh, Fei Xia, Andy Zeng, Wenhao Yu, Tingnan Zhang.

\subsection{Full task list}
\label{app:task lsit}

Here we show the list of tasks used in our evaluation as well as the instructions used for each task.

\begin{table}[ht]
\scriptsize
	\begin{center}
		\begin{tabular}{lll}
			\toprule
                \textbf{Task} & \textbf{Instructions} & \textbf{Expected Behavior}\\
                \midrule
                Facing sunrise & It's early in the morning, make the robot head towards the sun. & Robot face towards East. \\
                \midrule
                Facing sunset & It's late in the afternoon, make the robot head towards the sunset. & Robot face towards West. \\
                \midrule
                Sit down & Sit down low to ground with torso flat. & Robot's CoM drops lower and remain flat. \\
                \midrule
                Roll Over & I want the robot to roll by 180 degrees. & Robot's belly faces up.\\
                \midrule
                Spin & Spin fast. & Robot reach a fast turning speed.\\
                \midrule
                Lift one paw & I want the robot to lift its front right paw in the air. & The front right paw of the robot lifts up in the air.\\
                \midrule
                Lift paw higher & \makecell[l]{I want the robot to lift its front right paw in the air.\\Lift it even higher.} & The robot lifts its front right paw higher than before.\\
                \midrule
                Spin with lifted paws & \makecell[l]{Lift front left paw.\\Good, now lift diagonal paw as well.\\Good, in addition I want the robot to spin fast.}& Robot lifts front left and rear right paws while spin fast.\\
                \midrule
                Stand up on two feet & Make the robot stand upright on two back feet like a human. & Robot stands on two back feet and keep balance.\\
			\bottomrule
		\end{tabular}
	\end{center}
	\caption{List of tasks used in evaluation for the quadruped robot.}
	\label{tab:quadruped_tasks}
\end{table}

\begin{table}[ht]
\scriptsize
	\begin{center}
		\begin{tabular}{lll}
			\toprule
                \textbf{Task} & \textbf{Instructions} & \textbf{Expected Behavior}\\
                \midrule
                Touch object & Touch the \texttt{\{object\}}& Robot fingers in contact with the object. \\
                \midrule
                Lift object & Lift the \texttt{\{object\}} to \SI{0.5}{m} & The object needs to stay above \SI{0.4}{m} for 1s. \\
                \midrule
                Move object & Move the \texttt{\{object\_a\}} to \texttt{\{object\_b\}} & The distance between object needs to be smaller than \SI{0.1}{m}. \\
                \midrule
                Upright object & Place the \texttt{\{object\}} upright & The z axis of the object needs to be parallel to x-y plane. \\
                \midrule
                Flip object & Flip the \texttt{\{object\}} & The local up vector of the object should be pointing downward. \\
                \midrule
                Lift two objects & Lift the \texttt{\{object\_a\}} and \texttt{\{object\_b\}} at the same time. & Both objects need to stay above \SI{0.4}{m} for \SI{1}{s}. \\
                \midrule
                Turn on the faucet & Turn on the faucet. & The valve of the faucet needs to be turned 90 degrees. \\
                \midrule
                Open the drawer & Open the drawer. & The drawer needs to be pulled fully open.\\
			\bottomrule
		\end{tabular}
	\end{center}
	\caption{List of tasks used in evaluation for the dexterous manipulation.}
	\label{tab:manipulation_tasks}
\end{table}

\subsection{Baseline details}
\label{app:baseline_detail}
For the quadruped robot, we use the following three primitive skills:

\begin{itemize}
    \item \texttt{head\_towards(direction)} specifies a target heading direction \texttt{direction} for the robot to reach.
    \item \texttt{walk(forward\_speed, sideway\_speed, turning\_speed)} controls the robot to walk and turn in different directions. This is a common interface used in quadruped robots to navigate in different environments.
    \item \texttt{set\_joint\_poses(leg\_name, joint\_angles)} directly sets the joint positions for each DoF on the robot. To help the LLMs understand the joint angles, we provide a set of examples in the prompt.
\end{itemize}

For the dexterous manipulator robot, we use three primitive skills to control the robot motion and also a function to get access to the position of an object in the scene:

\begin{itemize}
  \item \texttt{end\_effector\_to(position)} moves the center of the robot hand's palm to the given \texttt{position}.
  \item \texttt{end\_effector\_open()} opens the hand of the robot by extending all fingers.
  \item \texttt{end\_effector\_close()} closes the hand to form a grasping pose.
  \item \texttt{get\_object\_position(obj\_name)} gets the position of a certain object in the scene.
  \item \texttt{get\_joint\_position(joint\_name)} gets the position of a certain joint in the scene.

\end{itemize}

\subsection{Additional illustrations for real-world results}

\begin{figure}[h]
    \centering
    \includegraphics[width=\textwidth]{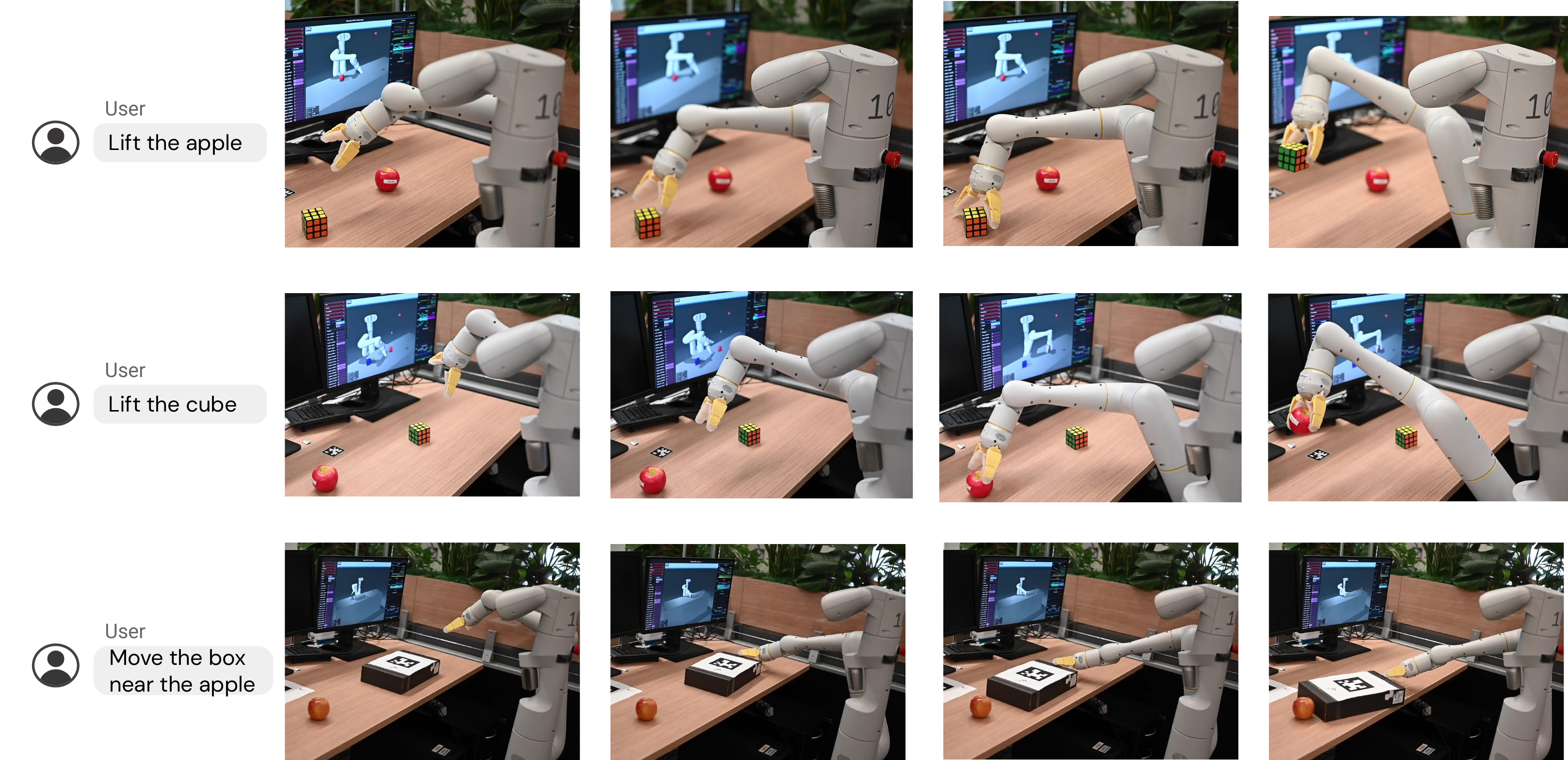}
    \caption{More illustrations for the real-world results for the proposed system.}
    \label{fig:l2r_real2}
\end{figure}

\subsection{Full Prompts}

\label{app:full_prompts}

Here we list the full prompts used in \emph{\langtorew} for all experiments used in this work.

\textbf{i) Motion Descriptor Prompt for Quadruped}

\begin{tcolorbox}[
    colframe=darkgray, 
    boxrule=0.2pt, 
    colback=lightgray!20, %
    arc=3pt, 
    fontupper=\small,
    breakable,
    halign=left
    ]

Describe the motion of a dog robot using the following form:\\

[start of description]

The torso of the robot should roll by [NUM: 0.0] degrees towards right, the torso should pitch upward at [NUM: 0.0] degrees.

The height of the robot's CoM or torso center should be at [NUM: 0.3] meters.

The robot should \{CHOICE: [face certain direction, turn at certain speed]\}. If facing certain direction, it should be facing \{CHOICE: [east, south, north, west]\}. If turning, it should turn at [NUM: 0.0] degrees/s.

The robot should \{CHOICE: [go to a certain location, move at certain speed]\}. If going to certain location, it should go to (x=[NUM: 0.0], y=[NUM: 0.0]). If moving at certain speed, it should move forward at [NUM: 0.0]m/s and sideways at [NUM: 0.0]m/s (positive means left).

[optional] front$\_$left foot lifted to [NUM: 0.0] meters high. 

[optional] back$\_$left foot lifted to [NUM: 0.0] meters high.

[optional] front$\_$right foot lifted to [NUM: 0.0] meters high.

[optional] back$\_$right foot lifted to [NUM: 0.0] meters high.

[optional] front$\_$left foot extend forward by [NUM: 0.0] meters.

[optional] back$\_$left foot extend forward by [NUM: 0.0] meters.

[optional] front$\_$right foot extend forward by [NUM: 0.0] meters.

[optional] back$\_$right foot extend forward by [NUM: 0.0] meters.

[optional] front$\_$left foot shifts inward laterally by [NUM: 0.0] meters.

[optional] back$\_$left foot shifts inward laterally by [NUM: 0.0] meters.

[optional] front$\_$right foot shifts inward laterally by [NUM: 0.0] meters.

[optional] back$\_$right foot shifts inward laterally by [NUM: 0.0] meters.

[optional] front$\_$left foot steps on the ground at a frequency of [NUM: 0.0] Hz, during the stepping motion, the foot will move [NUM: 0.0] meters up and down, and [NUM: 0.0] meters forward and back, drawing a circle as if it's walking \{CHOICE: forward, back\}, spending [NUM: 0.0] portion of the time in the air vs gait cycle.

[optional] back$\_$left foot steps on the ground at a frequency of [NUM: 0.0] Hz, during the stepping motion, the foot will move [NUM: 0.0] meters up and down, and [NUM: 0.0] meters forward and back, drawing a circle as if it's walking \{CHOICE: forward, back\}, spending [NUM: 0.0] portion of the time in the air vs gait cycle.

[optional] front$\_$right foot steps on the ground at a frequency of [NUM: 0.0] Hz, during the stepping motion, the foot will move [NUM: 0.0] meters up and down, and [NUM: 0.0] meters forward and back, drawing a circle as if it's walking \{CHOICE: forward, back\}, spending [NUM: 0.0] portion of the time in the air vs gait cycle.

[optional] back$\_$right foot steps on the ground at a frequency of [NUM: 0.0] Hz, during the stepping motion, the foot will move [NUM: 0.0] meters up and down, and [NUM: 0.0] meters forward and back, drawing a circle as if it's walking \{CHOICE: forward, back\}, spending [NUM: 0.0] portion of the time in the air vs gait cycle.

[optional] The phase offsets for the four legs should be front$\_$left: [NUM: 0.0], back$\_$left: [NUM: 0.0], front$\_$right: [NUM: 0.0], back$\_$right: [NUM: 0.0].

[end of description]
\\

Rules:

1. If you see phrases like [NUM: default$\_$value], replace the entire phrase with a numerical value.

2. If you see phrases like {CHOICE: [choice1, choice2, ...]}, it means you should replace the entire phrase with one of the choices listed. Be sure to replace all of them. If you are not sure about the value, just use your best judgement.

3. Phase offset is between [0, 1]. So if two legs' phase offset differs by 0 or 1 they are moving in synchronous. If they have phase offset difference of 0.5, they are moving opposite in the gait cycle.

4. The portion of air vs the gait cycle is between [0, 1]. So if it's 0, it means the foot will always stay on the ground, and if it's 1 it means the foot will always be in the air.

5. I will tell you a behavior/skill/task that I want the quadruped to perform and you will provide the full description of the quadruped motion, even if you may only need to change a few lines. Always start the description with [start of description] and end it with [end of description].

6. We can assume that the robot has a good low-level controller that maintains balance and stability as long as it's in a reasonable pose.

7. You can assume that the robot is capable of doing anything, even for the most challenging task.

8. The robot is about 0.3m high in CoM or torso center when it's standing on all four feet with horizontal body. It's about 0.65m high when it stand upright on two feet with vertical body. When the robot's torso/body is flat and parallel to the ground, the pitch and roll angles are both 0.

9. Holding a foot 0.0m in the air is the same as saying it should maintain contact with the ground.

10. Do not add additional descriptions not shown above. Only use the bullet points given in the template.

11. If a bullet point is marked [optional], do NOT add it unless it's absolutely needed.

12. Use as few bullet points as possible. Be concise.

\end{tcolorbox}

\textbf{ii) Reward Coder Prompt for Quadruped}

\begin{tcolorbox}[
    colframe=darkgray, 
    boxrule=0.2pt, 
    colback=lightgray!20, %
    arc=3pt, 
    fontupper=\small,
    breakable,
    halign=left
    ]

We have a description of a robot's motion and we want you to turn that into the corresponding program with following functions:

\begin{lstlisting}[breaklines=true]
def set_torso_targets(target_torso_height, target_torso_pitch, target_torso_roll, target_torso_location_xy, target_torso_velocity_xy, target_torso_heading, target_turning_speed)
\end{lstlisting}
target$\_$torso$\_$height: how high the torso wants to reach. When the robot is standing on all four feet in a normal standing pose, the torso is about 0.3m high.

target$\_$torso$\_$pitch: How much the torso should tilt up from a horizontal pose in radians. A positive number means robot is looking up, e.g. if the angle is 0.5*pi the robot will be looking upward, if the angel is 0, then robot will be looking forward.

target$\_$torso$\_$velocity$\_$xy: target torso moving velocity in local space, x is forward velocity, y is sideway velocity (positive means left).

target$\_$torso$\_$heading: the desired direction that the robot should face towards. The value of target$\_$torso$\_$heading is in the range of 0 to 2*pi, where 0 and 2*pi both mean East, pi being West, etc.

target$\_$turning$\_$speed: the desired turning speed of the torso in radians per second.

Remember: 
one of target$\_$torso$\_$location$\_$xy and target$\_$torso$\_$velocity$\_$xy must be None.
one of target$\_$torso$\_$heading and target$\_$turning$\_$speed must be None.
No other inputs can be None.
 
\begin{lstlisting}[breaklines=true]
def set_feet_pos_parameters(feet_name, lift_height, extend_forward, move_inward)
\end{lstlisting}
feet$\_$name is one of (``front$\_$left", ``back$\_$left", ``front$\_$right", ``back$\_$right").

lift$\_$height: how high should the foot be lifted in the air. If is None, disable this term. If it's set to 0, the foot will touch the ground.

extend$\_$forward: how much should the foot extend forward. If is None, disable this term.

move$\_$inward: how much should the foot move inward. If is None, disable this term.

\begin{lstlisting}[breaklines=true]
def set_feet_stepping_parameters(feet_name, stepping_frequency, air_ratio, phase_offset, swing_up_down, swing_forward_back, should_activate)
\end{lstlisting}
feet$\_$name is one of (``front$\_$left", ``rear$\_$left", ``front$\_$right", ``rear$\_$right").

air$\_$ratio (value from 0 to 1) describes how much time the foot spends in the air versus the whole gait cycle. If it's 0 the foot will always stay on ground, and if it's 1 it'll always stay in the air.

phase$\_$offset (value from 0 to 1) describes how the timing of the stepping motion differs between
different feet. For example, if the phase$\_$offset between two legs differs by 0.5, it means
one leg will start the stepping motion in the middle of the stepping motion cycle of the other leg.
swing$\_$up$\_$down is how much the foot swings vertical during the motion cycle.

swing$\_$forward$\_$back is how much the foot swings horizontally during the motion cycle.
If swing$\_$forward$\_$back is positive, the foot would look like it's going forward, if it's negative, the foot will look like it's going backward.

If should$\_$activate is False, the leg will not follow the stepping motion.

\begin{lstlisting}
def execute_plan(plan_duration=2)
\end{lstlisting}
This function sends the parameters to the robot and execute the plan for ``plan$\_$duration'' seconds, default to be 2

Example answer code:
\begin{lstlisting}[breaklines=true]
import numpy as np  # import numpy because we are using it below

reset_reward() # This is a new task so reset reward; otherwise we don't need it
set_torso_targets(0.1, np.deg2rad(5), np.deg2rad(15), (2, 3), None, None, np.deg2rad(10))

set_feet_pos_parameters("front_left", 0.1, 0.1, None)
set_feet_pos_parameters("back_left", None, None, 0.15)
set_feet_pos_parameters("front_right", None, None, None)
set_feet_pos_parameters("back_right", 0.0, 0.0, None)
set_feet_stepping_parameters("front_right", 2.0, 0.5, 0.2, 0.1, -0.05, True)
set_feet_stepping_parameters("back_left", 3.0, 0.7, 0.1, 0.1, 0.05, True)
set_feet_stepping_parameters("front_left", 0.0, 0.0, 0.0, 0.0, 0.0, False)
set_feet_stepping_parameters("back_right", 0.0, 0.0, 0.0, 0.0, 0.0, False)

execute_plan(4)
\end{lstlisting}

Remember:
1. Always format the code in code blocks.

2. Do not invent new functions or classes. The only allowed functions you can call are the ones listed above. Do not leave unimplemented code blocks in your response.

3. The only allowed library is numpy. Do not import or use any other library. If you use np, be sure to import numpy.

4. If you are not sure what value to use, just use your best judge. Do not use None for anything.

5. Do not calculate the position or direction of any object (except for the ones provided above). Just use a number directly based on your best guess.

6. For set$\_$torso$\_$targets, only the last four arguments (target$\_$torso$\_$location$\_$xy, target$\_$torso$\_$velocity$\_$xy, target$\_$torso$\_$heading, target$\_$turning$\_$speed) can be None. Do not set None for any other arguments.

7. Don't forget to call execute$\_$plan at the end.

\end{tcolorbox}

\textbf{iii) Baseline: Code-as-Policies Prompt for Quadruped}

\begin{tcolorbox}[
    colframe=darkgray, 
    boxrule=0.2pt, 
    colback=lightgray!20, %
    arc=3pt, 
    fontupper=\small,
    breakable,
    halign=left
    ]

We have a quadruped robot. It has 12 joints in total, three for each leg.
We can use the following functions to control its movements:

\begin{lstlisting}[breaklines=true]
def set_target_joint_angles(leg_name, target_joint_angles)
\end{lstlisting}
leg$\_$name is one of (``front$\_$left", ``back$\_$left", ``front$\_$right", ``back$\_$right").

target$\_$joint$\_$angles: a 3D vector that describes the target angle for the abduction/adduction, hip, and knee joint of the each leg.

\begin{lstlisting}
def walk(forward_speed, sideway_speed, turning_speed)
\end{lstlisting}
forward$\_$speed: how fast the robot should walk forward

sideway$\_$speed: how fast the robot should walk sideways

turning$\_$speed: how fast the robot should be turning (positive means turning right)

\begin{lstlisting}
def head_towards(heading_direction)
\end{lstlisting}
heading$\_$direction: target heading for the robot to reach, in the range of 0 to 2pi, where 0 means East, 0.5pi means North, pi means West, and 1.5pi means South.

\begin{lstlisting}
def execute_plan(plan_duration=10)
\end{lstlisting}
This function sends the parameters to the robot and execute the plan for ``plan$\_$duration" seconds, default to be 2
\\

Details about joint angles of each leg:
abduction/adduction joint controls the upper leg to swinging inward/outward. 
When it's positive, legs will swing outward (swing to the right for right legs and left for left legs).
When it's negative, legs will swing inward.

hip joint controls the upper leg to rotate around the shoulder.
When it's zero, the upper leg is parallel to the torso (hip is same height as shoulder), pointing backward.
When it's positive, the upper leg rotates downward so the knee is below the shoulder. When it's 0.5pi, it's perpendicular to the torso, pointing downward.
When it's negative, the upper leg rotates upward so the knee is higher than the shoulder.

knee joint controls the lower leg to rotate around the knee.
When it's zero, the lower leg is folded closer to the upper leg.
knee joint angle can only be positive. When it's 0.5pi, the lower leg is perpendicular to the upper leg. When it's pi, the lower leg is fully streching out and parallel to the upper leg.

Here are a few examples for setting the joint angles to make the robot reach a few key poses:
standing on all four feet:

\begin{lstlisting}
set_target_joint_angles("front_left", [0, 1, 1.5])
set_target_joint_angles("back_left", [0, 0.75, 1.5])
set_target_joint_angles("front_right", [0, 1, 1.5])
set_target_joint_angles("back_right", [0, 0.75, 1.5])
execute_plan()
\end{lstlisting}

sit down on the floor:
\begin{lstlisting}
set_target_joint_angles("front_left", [0, 0, 0])
set_target_joint_angles("back_left", [0, 0, 0])
set_target_joint_angles("front_right", [0, 0, 0])
set_target_joint_angles("back_right", [0, 0, 0])
execute_plan()
\end{lstlisting}

lift front left foot:
\begin{lstlisting}
set_target_joint_angles("front_left", [0, 0.45, 0.35])
set_target_joint_angles("back_left", [0, 1, 1.5])
set_target_joint_angles("front_right", [0, 1.4, 1.5])
set_target_joint_angles("back_right", [0, 1, 1.5])
execute_plan()
\end{lstlisting}

lift back left foot:
\begin{lstlisting}
set_target_joint_angles("front_left", [0, 0.5, 1.5])
set_target_joint_angles("back_left", [0, 0.45, 0.35])
set_target_joint_angles("front_right", [0, 0.5, 1.5])
set_target_joint_angles("back_right", [0, 0.5, 1.5])
execute_plan()
\end{lstlisting}

Remember:

1. Always start your response with [start analysis]. Provide your analysis of the problem within 100 words, then end it with [end analysis].

2. After analysis, start your code response, format the code in code blocks.

3. Do not invent new functions or classes. The only allowed functions you can call are the ones listed above. Do not leave unimplemented code blocks in your response.

4. The only allowed library is numpy. Do not import or use any other library. If you use np, be sure to import numpy.

5. If you are not sure what value to use, just use your best judge. Do not use None for anything.

6. Do not calculate the position or direction of any object (except for the ones provided above). Just use a number directly based on your best guess.

7. Write the code as concisely as possible and try not to define additional variables.

8. If you define a new function for the skill, be sure to call it somewhere.

9. Be sure to call execute$\_$plan at the end.

\end{tcolorbox}

\textbf{iv) Motion Descriptor Prompt for Dexterous Manipulator}

\begin{tcolorbox}[
    colframe=darkgray, 
    boxrule=0.2pt, 
    colback=lightgray!20, %
    arc=3pt, 
    fontupper=\small,
    breakable,
    halign=left,
    ]

We have a dexterous manipulator and we want you to help plan how it should move to perform tasks using the following template:
\\

[start of description]

To perform this task, the manipulator's palm should move close to \{CHOICE: apple, banana, box, bowl, drawer$\_$handle, faucet$\_$handle, drawer$\_$center, rest$\_$position\}.

object1=\{CHOICE: apple, banana, box, bowl, drawer$\_$handle, faucet$\_$handle, drawer$\_$center\} should be close to object2=\{CHOICE: apple, banana, box, bowl, drawer$\_$handle, faucet$\_$handle, drawer$\_$center, nothing\}.

[optional] object1 needs to be rotated by [NUM: 0.0] degrees along x axis.

[optional] object2 needs to be rotated by [NUM: 0.0] degrees along x axis.

[optional] object1 needs to be lifted to a height of [NUM: 0.0]m at the end.

[optional] object2 needs to be lifted to a height of [NUM: 0.0]m at the end.

[optional] object3=\{CHOICE: drawer, faucet\} needs to be \{CHOICE: open, closed\}.


[end of description]
\\

Rules:

1. If you see phrases like [NUM: default$\_$value], replace the entire phrase with a numerical value.

2. If you see phrases like \{CHOICE: choice1, choice2, ...\}, it means you should replace the entire phrase with one of the choices listed.

3. If you see [optional], it means you only add that line if necessary for the task, otherwise remove that line.

4. The environment contains apple, banana, box, bowl, drawer$\_$handle, faucet$\_$handle. Do not invent new objects not listed here.

5. The bowl is large enough to have all other object put in there.

6. I will tell you a behavior/skill/task that I want the manipulator to perform and you will provide the full plan, even if you may only need to change a few lines. Always start the description with [start of plan] and end it with [end of plan].

7. You can assume that the robot is capable of doing anything, even for the most challenging task.

8. Your plan should be as close to the provided template as possible. Do not include additional details.

\end{tcolorbox}

\textbf{v) Reward Coder Prompt for Dexterous Manipulator}

\begin{tcolorbox}[
    colframe=darkgray, 
    boxrule=0.2pt, 
    colback=lightgray!20, %
    arc=3pt, 
    fontupper=\small,
    breakable,
    halign=left
    ]

We have a plan of a robot arm with palm to manipulate objects and we want you to turn that into the corresponding program with following functions:

\begin{lstlisting}
def set_l2_distance_reward(name_obj_A, name_obj_B)
\end{lstlisting}
where name$\_$obj$\_$A and name$\_$obj$\_$B are selected from [``palm", ``apple", ``banana", ``box", ``bowl", ``drawer$\_$handle", ``faucet$\_$handle", ``drawer$\_$center", ``rest$\_$position"].
This term sets a reward for minimizing l2 distance between name$\_$obj$\_$A and name$\_$obj$\_$B so they get closer to each other.
rest$\_$position is the default position for the palm when it's holding in the air.

\begin{lstlisting}
def set_obj_orientation_reward(name_obj, x_axis_rotation_radians)
\end{lstlisting}
this term encourages the orientation of name$\_$obj to be close to the target (specified by x$\_$axis$\_$rotation$\_$radians).

\begin{lstlisting}
def execute_plan(duration=2)
\end{lstlisting}
This function sends the parameters to the robot and execute the plan for ``duration'' seconds, default to be 2.

\begin{lstlisting}
def set_joint_fraction_reward(name_joint, fraction)
\end{lstlisting}
This function sets the joint to a certain value between 0 and 1. 0 means close and 1 means open.
name$\_$joint needs to be select from ['drawer', 'faucet']

\begin{lstlisting}
def set_obj_z_position_reward(name_obj, z_height)
\end{lstlisting}
this term encourages the orientation of name\_obj to be close to the height (specified by z$\_$height).

\begin{lstlisting}
def reset_reward()
\end{lstlisting}
This function resets the reward to default values.

Example plan:
To perform this task, the manipulator's palm should move close to object1=apple.
object1 should be close to object2=bowl.
object2 needs to be rotated by 30 degrees along x axis.
object2 needs to be lifted to a height of 1.0.

This is the first plan for a new task.

Example answer code:
\begin{lstlisting}
import numpy as np

reset_reward() # This is a new task so reset reward; otherwise we don't need it
set_l2_distance_reward("palm", "apple")
set_l2_distance_reward("apple", "bowl")
set_obj_orientation_reward("bowl", np.deg2rad(30))
set_obj_z_position_reward("bowl", 1.0)

execute_plan(4)
\end{lstlisting}

Remember:

1. Always format the code in code blocks. In your response execute$\_$plan should be called exactly once at the end.

2. Do not invent new functions or classes. The only allowed functions you can call are the ones listed above. Do not leave unimplemented code blocks in your response.

3. The only allowed library is numpy. Do not import or use any other library.

4. If you are not sure what value to use, just use your best judge. Do not use None for anything.

5. Do not calculate the position or direction of any object (except for the ones provided above). Just use a number directly based on your best guess.

6. You do not need to make the robot do extra things not mentioned in the plan such as stopping the robot.

\end{tcolorbox}

\textbf{vi) Baseline: Code-as-Policies Prompt for Dexterous Manipulator}

\begin{tcolorbox}[
    colframe=darkgray, 
    boxrule=0.2pt, 
    colback=lightgray!20, %
    arc=3pt, 
    fontupper=\small,
    breakable,
    halign=left
    ]

We have a manipulator and we want you to help plan how it should move to perform tasks using the following APIs:

\begin{lstlisting}
def end_effector_to(position_obj)
\end{lstlisting}
position$\_$obj is a list of 3 float numbers [x,y,z]

\begin{lstlisting}
def end_effector_open()
\end{lstlisting}
Open the end effector.

\begin{lstlisting}
def end_effector_close()
\end{lstlisting}

Close the end effector.

\begin{lstlisting}
def get_object_position(obj_name)
\end{lstlisting}
Given an object name, return a list of 3 float numbers [x,y,z] for the object position.
the object can come from a list of [``apple", ``banana", ``bowl", ``box", ``drawer$\_$handle", ``faucet$\_$handle", ``drawer$\_$center", ``rest$\_$position"]

\begin{lstlisting}
def get_normalized_joint_position(joint_name)
\end{lstlisting}

Given an joint name, return a float numbers x.
the joint can come from a list of [``drawer", ``faucet"]

\begin{lstlisting}
def reset()
\end{lstlisting}
Reset the agent.

Example answer code:
\begin{lstlisting}
import numpy as np

reset()
apple_pos = get_object_position("apple")
end_effector_to(apple$_pos)
\end{lstlisting}

Remember:

1. Always format the code in code blocks.

2. Do not invent new functions or classes. The only allowed functions you can call are the ones listed above. Do not leave unimplemented code blocks in your response.

3. The only allowed library is numpy. Do not import or use any other library.

4. If you are not sure what value to use, just use your best judge. Do not use None for anything.

5. You do not need to make the robot do extra things not mentioned in the plan such as stopping the robot.

6. Try your best to generate code despite the lack of context.

\end{tcolorbox}





\subsection{Reward functions used in our experiments}
\label{app:reward_terms}

In this work we use a set of generic reward functions for each embodiment that the LLMs can modulate. More specifically, we design a set of residual terms as in Equation \ref{eq:mjpc_objective} that are optimized to reach zero by internally converting them to a l2 loss. Thus given a residual term $r(\cdot)$ a reward term can be recovered by $-||r(\cdot)||_2^2$. Below we describe the full set of residual terms we use in our experiments for each embodiment. For each term we select the weights for them to have about the same magnitude. The reward coder can adjust the parameters in each term and optionally set the weight to zero to disable a term.

\subsubsection{Quadruped}

Table \ref{tab:quadruped_rewards} shows the residual terms used in the quadruped tasks. Note that for the foot-related terms, they are repeated for all four feet respectively. Furthermore, LLMs can optionally set the target foot positions $\bar{\mathbf{fp}}$ directly or through a periodic function $max(a sin(b 2\pi+c), 0)$ where $a$ is the magnitude of the motion, $b$ is the frequency, and $c$ is the phase offset.

\begin{table}[h]
\scriptsize
	\begin{center}
		\begin{tabular}{lll}
			\toprule
                \textbf{Residual Term} & \textbf{Formulation} & \textbf{Default weight}\\
                CoM X-Y position & $|\mathbf{p}_{xy}-\bar{\mathbf{p}_{xy}}|$ & 0.3 \\
                CoM height & $\mathbf{p}_{z}-\bar{\mathbf{p}_{z}}$ & 1.0 \\
                base yaw & $\mathbf{p}_{yaw}-\bar{\mathbf{p}_{yaw}}$ & 0.3 \\
                base pitch & $\mathbf{p}_{pitch}-\bar{\mathbf{p}_{pitch}}$ & 0.6 \\
                base roll & $\mathbf{p}_{roll}-\bar{\mathbf{p}_{roll}}$ &  0.1\\
                forward velocity & $\mathbf{\dot{p}}_x - \bar{\mathbf{\dot{p}}_x}$ & 0.1 \\
                sideways velocity & $\mathbf{\dot{p}}_y - \bar{\mathbf{\dot{p}}_y}$ & 0.1 \\
                yaw speed & $\mathbf{\dot{p}}_{yaw} - \bar{\mathbf{\dot{p}}_{yaw}}$ & 0.1 \\
                foot local position x & $\mathbf{fp}_x - \bar{\mathbf{fp}_x}$ & 1  \\
                foot local position y & $\mathbf{fp}_y - \bar{\mathbf{fp}_y}$ & 1 \\
                foot local position z & $\mathbf{fp}_z - \bar{\mathbf{fp}_z}$ & 2 \\
			\bottomrule
		\end{tabular}
	\end{center}
	\caption{List of residual terms used for the quadruped robot. $\mathbf{p}$ denotes the position and orientation of the robot's torso. $\mathbf{fp}$ denotes the position of the robot's foot (in local space). $\bar{(\cdot)}$ means the target value and $\dot{(\cdot)}$ means the time-derivative of the quantity.}
	\label{tab:quadruped_rewards}
\end{table}

\subsubsection{Dexterous Manipulator}

\begin{table}[ht]
\scriptsize
	\begin{center}
		\begin{tabular}{lll}
			\toprule
                \textbf{Residual Term} & \textbf{Formulation} & \textbf{Default weight}\\
                move obj1 close to obj2 & $|\mathbf{c1}_{xyz} - \mathbf{c2}_{xyz}|$ & 5 \\
                move obj to target X-Y position & $|\mathbf{c}_{z} - \bar{\mathbf{c}}_z|$ & 5 \\
                move obj to target height & $|\mathbf{c}_{xy} - \bar{\mathbf{c}}_{xy}|$ & 10 \\
                move obj to target orientation & $|\mathbf{o}_{obj} - \bar{\mathbf{o}}|$ & \\
                move joint to target value & $q-\bar{q}$ & 10 \\
			\bottomrule
		\end{tabular}
	\end{center}
	\caption{List of reward terms used for the dexterous manipulator robot. $\mathbf{c}$ denotes the position of the object, $\mathbf{o}$ denotes the orientation of the object, $q$ is the degree of freedom to be manipulated.}
	\label{tab:manipulation_rewards}
\end{table}

\subsubsection{Sim-to-Real residual term}
\label{app:sim2real_term}

As seen in the supplementary video, MuJoCo MPC can discover highly dynamic and dexterous manipulation skills that exceeds the capabilities of existing hardwares. To enable successful deployment on the hardware, we design a regularization term to help achieve steady motions on the robot. Specifically, we use the following residual term:

\begin{align*}
r_{sim2real}=&
3\begin{cases}
    \dot{\mathbf{p}_{ee}},& \text{if } ||\dot{\mathbf{p}_{ee}}|| > 0.3\\
    0,              & \text{otherwise}
\end{cases}
\\&+
\begin{cases}
    \dot{\mathbf{q}},& \text{if } ||\dot{\mathbf{q}}|| > 0.7\\
    0,              & \text{otherwise}
\end{cases}
\\&+
0.05\dot{\mathbf{p}_{obj}}
\\&+
0.1\begin{cases}
    \dot{\mathbf{p}_{ee}}-\dot{\mathbf{p}_{obj}},& \text{if } ||\mathbf{p}_{ee}-\mathbf{p}_{obj}|| < 0.1\\
    0,              & \text{otherwise}
\end{cases}
\\&+
0.4\begin{cases}
    \mathbf{q}_{gripper}-1.0,& \text{if } ||\mathbf{p}_{ee}-\mathbf{p}_{obj}|| > 0.1\\
    \mathbf{q}_{gripper},              & \text{otherwise}
\end{cases}
,
\end{align*}
where $\dot{\mathbf{p}_{ee}}$ is the robot end effector velocity, $\dot{\mathbf{q}}$ is the robot joint velocity, $\dot{\mathbf{p}_{obj}}$ is the object velocity. The first two terms regularizes the joint and end effector velocities to encourage them to stay at a low-speed region to prevent jerky robot motions. The third and fourth term encourage objects to move slowly and match gripper speed. This is to discourage dynamic in-hand movements. The last term encourages gripper to open while being away from objects and close when approaching objects. This term helps improve the grasping skill transfer by encouraging a more clean grasp motion.

\subsection{Detailed evaluation results for each task}

\begin{figure}[ht]
    \centering
    \includegraphics[width=\textwidth]{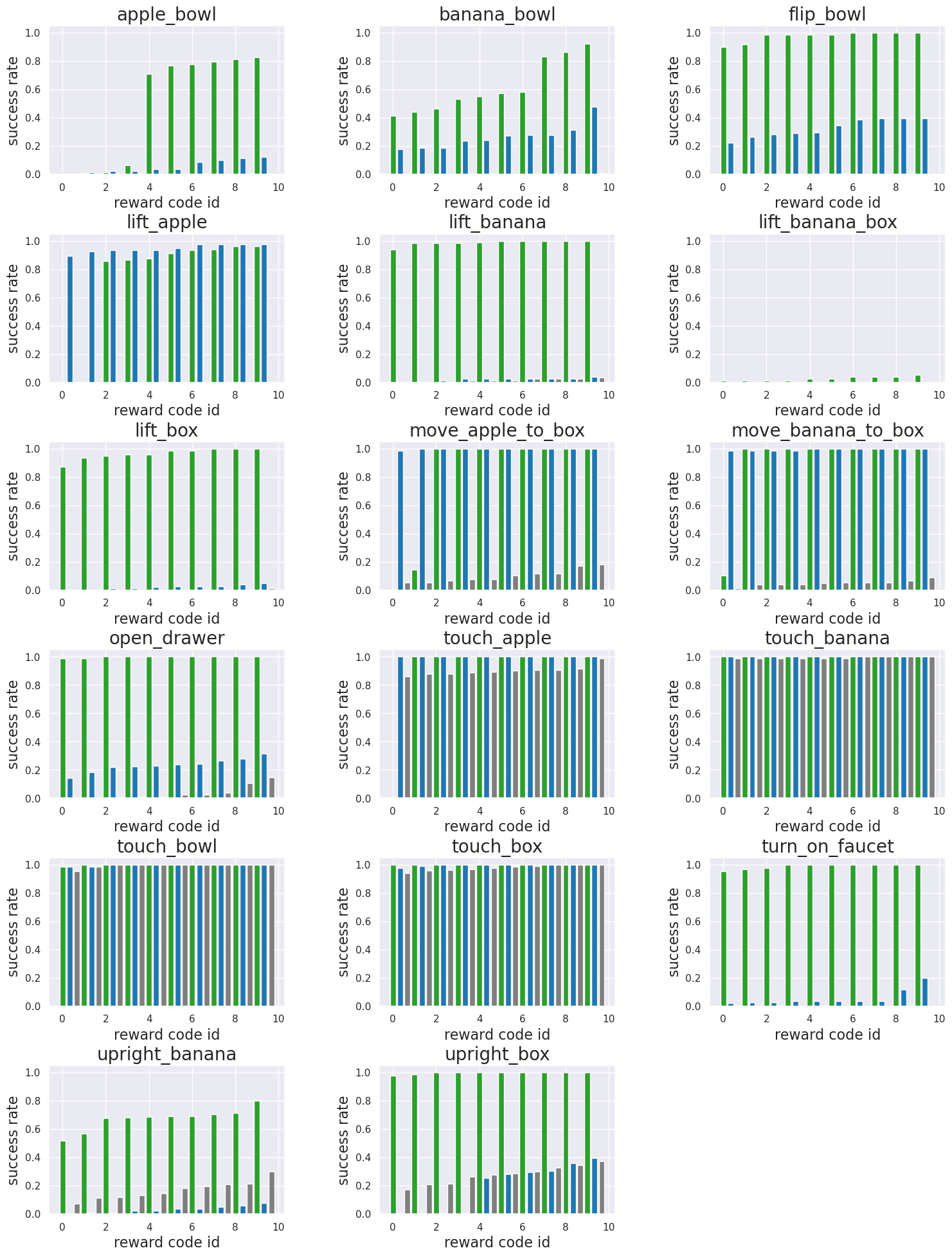}
    \\
    \includegraphics[width=0.6\textwidth, center]{figures/legend.png}
    \caption{Full evaluation results for the Dexterous Manipulator robot. Note that to better interpret the results, we order the generated reward code in the figures based on the mean success rate.}
    \label{fig:manipulation_stat}
\end{figure}

\begin{figure}[ht]
    \centering
    \includegraphics[width=\textwidth]{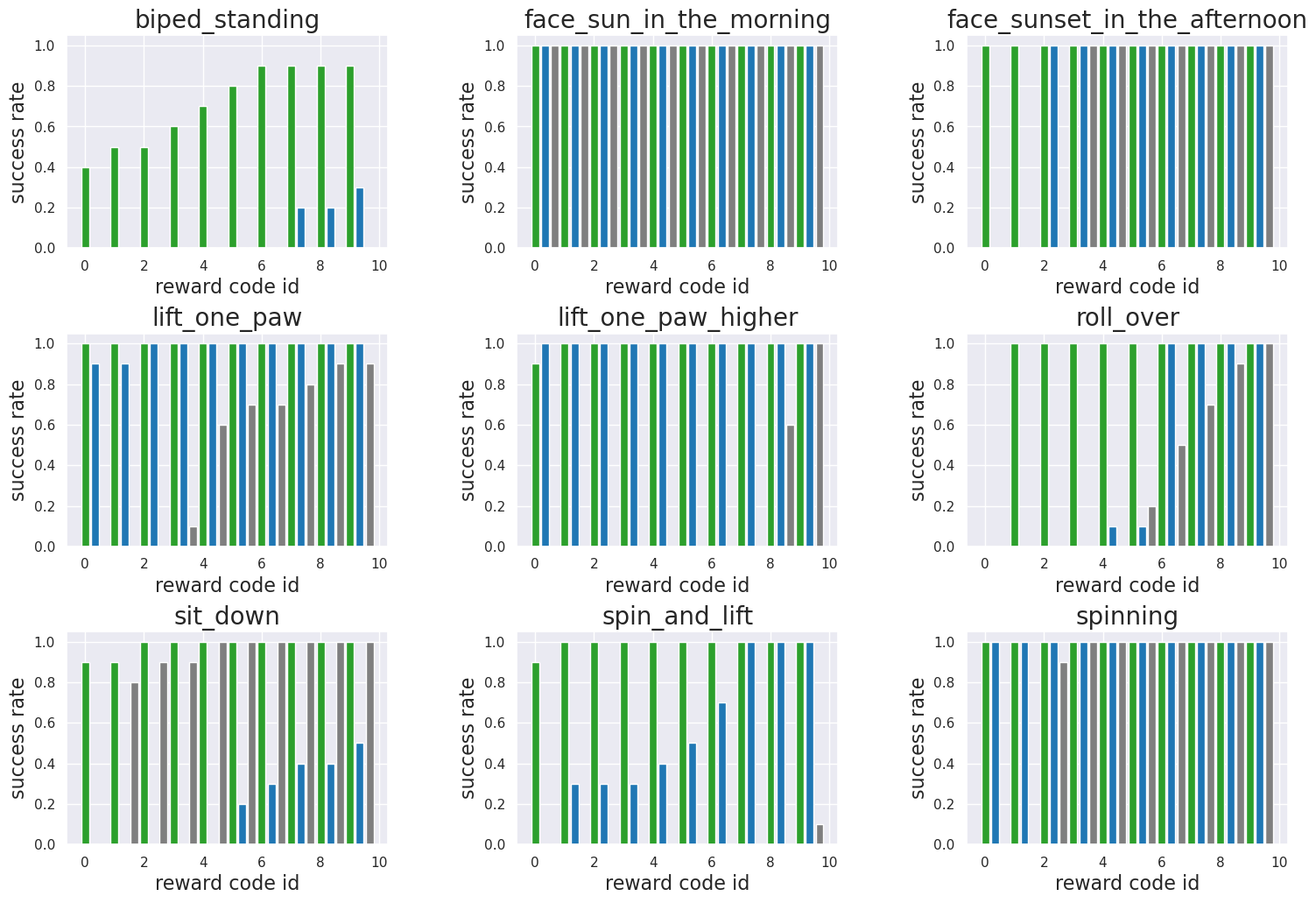}
    \\
    \includegraphics[width=0.6\textwidth, center]{figures/legend.png}
    \caption{Full evaluation results for the Quadruped robot. Note that to better interpret the results, we order the generated reward code in the figures based on the mean success rate.}
    \label{fig:quadruped_stat}
\end{figure}

Figure \ref{fig:quadruped_stat} and Figure \ref{fig:manipulation_stat} shows the full evaluation results for individual tasks in \cref{ssec:eval}. For each task, we query the \emph{\langtorew} LLM 10 times. For each generated reward code, we evaluate on MJPC for 50 times and report the success rate. Therefore the results here shows the individual performance of \emph{\langtorew} and \emph{\rewtoact}. We can see that for most tasks \emph{\rewtoact} achieves near perfect success rate as long as the right reward is provided, while for tasks that require more dynamic motion or more precise manipulation (e.g. biped standing, and upright banana) \emph{\rewtoact} shows more variance in performance.

\subsection{Example outputs}
\label{app:example_outputs}

To help understand the behavior of different methods discussed in this work, here we show sampled responses for two tasks across different methods: \emph{Biped standing} for quadruped, and \emph{Turn on faucet} for dexterous manipulator. For each method, we sample $2$-$3$ responses to show both successful and failed results.

\subsubsection{Sampled results for \emph{Biped standing} task}

\textbf{User instruction}: \emph{Make the robot stand upright on two back feet like a human.}

\textbf{Ours}

\noindent\begin{tcolorbox}[
    colframe=darkgray, 
    boxrule=0.5pt, 
    colback=green!5, %
    arc=3pt, 
    fontupper=\small,
    nobeforeafter, title=Sample 1,
    ]

\begin{lstlisting}[
    breaklines=true
]
import numpy as np

reset_reward()
set_torso_targets(0.65, np.deg2rad(90), np.deg2rad(0), (0.0, 0.0), None, 0, None)

set_feet_pos_parameters("front_left", 0.65, None, None)
set_feet_pos_parameters("back_left", 0.0, None, None)
set_feet_pos_parameters("front_right", 0.65, None, None)
set_feet_pos_parameters("back_right", 0.0, None, None)

execute_plan()

\end{lstlisting}

\end{tcolorbox}

\begin{tcolorbox}[
    colframe=darkgray, 
    boxrule=0.5pt, 
    colback=red!5, %
    arc=3pt, 
    fontupper=\small,
    nobeforeafter, title=Sample 2,
    ]

\begin{lstlisting}[
    breaklines=true
]
import numpy as np

reset_reward()
set_torso_targets(0.65, np.deg2rad(90), np.deg2rad(0), (0.0, 0.0), None, 0, None)

set_feet_pos_parameters("front_left", 0.65, None, None)
set_feet_pos_parameters("back_left", 0.0, None, None)
set_feet_pos_parameters("front_right", 0.65, None, None)
set_feet_pos_parameters("back_right", 0.0, None, None)
\end{lstlisting}

\end{tcolorbox}

\begin{tcolorbox}[
    colframe=darkgray, 
    boxrule=0.5pt, 
    colback=green!5, %
    arc=3pt, 
    fontupper=\small,
    width=\textwidth, nobeforeafter, title=Sample 3,
    ]

\begin{lstlisting}[
    breaklines=true
]
import numpy as np

reset_reward()
set_torso_targets(0.65, np.deg2rad(90), np.deg2rad(0), (0, 0), None, 0, None)

set_feet_pos_parameters("front_left", 0.65, None, None)
set_feet_pos_parameters("back_left", 0.0, None, None)
set_feet_pos_parameters("front_right", 0.65, None, None)
set_feet_pos_parameters("back_right", 0.0, None, None)

set_feet_stepping_parameters("front_left", 0.0, 0.0, 0.0, 0.0, 0.0, False)
set_feet_stepping_parameters("back_left", 0.0, 0.0, 0.0, 0.0, 0.0, False)
set_feet_stepping_parameters("front_right", 0.0, 0.0, 0.0, 0.0, 0.0, False)
set_feet_stepping_parameters("back_right", 0.0, 0.0, 0.0, 0.0, 0.0, False)

execute_plan()
\end{lstlisting}

\end{tcolorbox}

\textbf{Reward Coder only}

\noindent\begin{tcolorbox}[
    colframe=darkgray, 
    boxrule=0.5pt, 
    colback=red!5, %
    arc=3pt, 
    fontupper=\small,
    nobeforeafter, title=Sample 1,
    ]

\begin{lstlisting}[
    breaklines=true
]
import numpy as np

# Set torso targets to stand on back feet
set_torso_targets(target_torso_height=0.6, target_torso_pitch=0.5*np.pi, target_torso_roll=0.0, target_torso_location_xy=None, target_torso_velocity_xy=None, target_torso_heading=None, target_turning_speed=None)

# Lift front feet off the ground
set_feet_pos_parameters(feet_name="front_left", lift_height=0.3, extend_forward=None, move_inward=None)
set_feet_pos_parameters(feet_name="front_right", lift_height=0.3, extend_forward=None, move_inward=None)

# Execute the plan
execute_plan(plan_duration=3)

\end{lstlisting}

\end{tcolorbox}

\begin{tcolorbox}[
    colframe=darkgray, 
    boxrule=0.5pt, 
    colback=red!5, %
    arc=3pt, 
    fontupper=\small,
    nobeforeafter, title=Sample 2,
    ]

\begin{lstlisting}[
    breaklines=true
]

import numpy as np

# Set the torso targets to stand on two back feet
set_torso_targets(target_torso_height=0.6, target_torso_pitch=0.0, target_torso_roll=0.0, target_torso_location_xy=None, target_torso_velocity_xy=(0, 0), target_torso_heading=None, target_turning_speed=None)

# Lift the front feet off the ground
set_feet_pos_parameters("front_left", lift_height=0.3, extend_forward=None, move_inward=None)
set_feet_pos_parameters('front_right', lift_height=0.3, extend_forward=None, move_inward=None)

# Execute the plan
execute_plan(plan_duration=3)
\end{lstlisting}

\end{tcolorbox}

\textbf{Code-as-Policies}

\noindent\begin{tcolorbox}[
    colframe=darkgray, 
    boxrule=0.5pt, 
    colback=green!5, %
    arc=3pt, 
    fontupper=\small,
    nobeforeafter, title=Sample 1,
    ]

\begin{lstlisting}[
    breaklines=true
]
set_target_joint_angles("front_left", [0, 1, 1.5])
set_target_joint_angles("back_left", [0, 0.5, 0])
set_target_joint_angles("front_right", [0, 1, 1.5])
set_target_joint_angles("back_right", [0, 0.5, 0])
execute_plan()

\end{lstlisting}

\end{tcolorbox}

\begin{tcolorbox}[
    colframe=darkgray, 
    boxrule=0.5pt, 
    colback=red!5, %
    arc=3pt, 
    fontupper=\small,
    nobeforeafter, title=Sample 2,
    ]

\begin{lstlisting}[
    breaklines=true
]
set_target_joint_angles("front_left", [0, 0.5, 1.5])
set_target_joint_angles("back_left", [0, 1.2, 0.5])
set_target_joint_angles("front_right", [0, 0.5, 1.5])
set_target_joint_angles("back_right", [0, 1.2, 0.5])
execute_plan()
\end{lstlisting}

\end{tcolorbox}

\begin{tcolorbox}[
    colframe=darkgray, 
    boxrule=0.5pt, 
    colback=green!5, %
    arc=3pt, 
    fontupper=\small,
    width=\textwidth, nobeforeafter, title=Sample 3,
    ]

\begin{lstlisting}[
    breaklines=true
]
import numpy as np

def set_legs_to_pose(pose):
    set_target_joint_angles("front_left", pose[0])
    set_target_joint_angles("back_left", pose[1])
    set_target_joint_angles("front_right", pose[2])
    set_target_joint_angles("back_right", pose[3])

def execute_pose(pose, duration=2):
    set_legs_to_pose(pose)
    execute_plan(duration)

# set initial pose
initial_pose = [
    [0, 0, 0], # front_left
    [0, 0, 0], # back_left
    [0, 0, 0], # front_right
    [0, 0, 0] # back_right
]
set_legs_to_pose(initial_pose)
execute_plan()

# set target pose
target_pose = [
    [0, 0.3, 0.5], # front_left
    [0, 1.2, 1.5], # back_left
    [0, 0, 0], # front_right
    [0, 1.2, 1.5] # back_right
]
execute_pose(target_pose)
\end{lstlisting}

\end{tcolorbox}

\subsubsection{Sampled results for \emph{Turn on faucet} task}

\textbf{User instruction}: \emph{Turn on faucet.}

\textbf{Ours}

\noindent\begin{tcolorbox}[
    colframe=darkgray, 
    boxrule=0.5pt, 
    colback=green!5, %
    arc=3pt, 
    fontupper=\small,
    nobeforeafter, title=Sample 1,
    ]

\begin{lstlisting}[
    breaklines=true
]
import numpy as np

reset_reward() # This is a new task so reset reward; otherwise we don't need it
set_l2_distance_reward("palm", "faucet_handle")
set_obj_orientation_reward("faucet_handle", np.deg2rad(90))
set_joint_fraction_reward("faucet", 1)

execute_plan(2)

\end{lstlisting}

\end{tcolorbox}

\begin{tcolorbox}[
    colframe=darkgray, 
    boxrule=0.5pt, 
    colback=green!5, %
    arc=3pt, 
    fontupper=\small,
    nobeforeafter, title=Sample 2,
    ]

\begin{lstlisting}[
    breaklines=true
]
import numpy as np

reset_reward() # This is a new task so reset reward; otherwise we don't need it
set_l2_distance_reward("palm", "faucet_handle")
set_joint_fraction_reward("faucet", 1) # 1 means open

execute_plan(2)
\end{lstlisting}

\end{tcolorbox}

\textbf{Reward Coder only}

\noindent\begin{tcolorbox}[
    colframe=darkgray, 
    boxrule=0.5pt, 
    colback=red!5, %
    arc=3pt, 
    fontupper=\small,
    nobeforeafter, title=Sample 1,
    ]

\begin{lstlisting}[
    breaklines=true,
]
import numpy as np

reset_reward()
set_joint_fraction_reward("faucet", 1)

execute_plan()
\end{lstlisting}

\end{tcolorbox}

\begin{tcolorbox}[
    colframe=darkgray, 
    boxrule=0.5pt, 
    colback=red!5, %
    arc=3pt, 
    fontupper=\small,
    nobeforeafter, title=Sample 2,
    ]

\begin{lstlisting}[
    breaklines=true
]
import numpy as np

reset_reward() # Reset reward for a new task
set_joint_fraction_reward("faucet", 1) # Turn on the faucet by setting the joint fraction to 1 (open)

execute_plan(2)
\end{lstlisting}

\end{tcolorbox}

\textbf{Code-as-Policies}

\noindent\begin{tcolorbox}[
    colframe=darkgray, 
    boxrule=0.5pt, 
    colback=red!5, %
    arc=3pt, 
    fontupper=\small,
    nobeforeafter, title=Sample 1,
    ]

\begin{lstlisting}[
    breaklines=true
]
reset()
faucet_handle_position = get_object_position("faucet_handle")
end_effector_to(faucet_handle_position)
end_effector_close()
normalized_joint_position = get_normalized_joint_position("faucet")
end_effector_open()

\end{lstlisting}

\end{tcolorbox}

\begin{tcolorbox}[
    colframe=darkgray, 
    boxrule=0.5pt, 
    colback=red!5, %
    arc=3pt, 
    fontupper=\small,
    nobeforeafter, title=Sample 2,
    ]

\begin{lstlisting}[
    breaklines=true
]
#python
import numpy as np

reset()
faucet_handle_pos = get_object_position("faucet_handle")
end_effector_to(faucet_handle_pos)
end_effector_open()
end_effector_close()

normalized_faucet_position = get_normalized_joint_position("faucet")
target_faucet_position = np.clip(normalized_faucet_position + 0.5, 0, 1)

end_effector_to([faucet_handle_pos[0], faucet_handle_pos[1], faucet_handle_pos[2] + 0.1])
end_effector_to([faucet_handle_pos[0], faucet_handle_pos[1], faucet_handle_pos[2]])
end_effector_open()
\end{lstlisting}

\end{tcolorbox}

\subsection{Full outputs for the interactive examples}
\label{app:interact_code_output}

Below we show the output reward code from our system for the two interactive examples: tasking a quadruped robot to perform moonwalk, and tasking a dexterous manipulator robot to place an apple in a drawer.

\textbf{Quadruped moonwalk}

\noindent\begin{tcolorbox}[
    colframe=darkgray, 
    boxrule=0.5pt, 
    colback=lightgray!20, %
    arc=3pt, 
    fontupper=\small,
    nobeforeafter, title={Instruction 1: Make the robot stand upright on two back feet like a human.},
    ]

\begin{lstlisting}[
    breaklines=true
]
import numpy as np

reset_reward()
set_torso_targets(0.65, np.deg2rad(90), np.deg2rad(0), (0.0, 0.0), None, 0, None)

set_feet_pos_parameters("front_left", 0.65, None, None)
set_feet_pos_parameters("back_left", 0.0, None, None)
set_feet_pos_parameters("front_right", 0.65, None, None)
set_feet_pos_parameters("back_right", 0.0, None, None)

execute_plan()

\end{lstlisting}

\end{tcolorbox}

\begin{tcolorbox}[
    colframe=darkgray, 
    boxrule=0.5pt, 
    colback=lightgray!20, %
    arc=3pt, 
    fontupper=\small,
    nobeforeafter, title={Instruction 2: Good, you actually don't need to keep the front paws at certain height, just leave it to the controller.},
    ]

\begin{lstlisting}[
    breaklines=true,
    escapechar=!
]

import numpy as np

reset_reward()
set_torso_targets(0.65, np.deg2rad(90), np.deg2rad(0), (0.0, 0.0), None, 0, None)

set_feet_pos_parameters("back_left", 0.0, None, None)
set_feet_pos_parameters("back_right", 0.0, None, None)

!\colorbox{lightgreen}{\# Set stepping parameters for all feet to disable stepping motion}!
!\colorbox{lightgreen}{for foot in ["front$\_$left", "rear$\_$left", "front$\_$right", "rear$\_$right"]:}!
!\colorbox{lightgreen}{   \  set$\_$feet$\_$stepping$\_$parameters(foot, 0.0, 0.0, 0.0, 0.0, 0.0, False)}!


execute_plan()

\end{lstlisting}

\end{tcolorbox}

\begin{tcolorbox}[
    colframe=darkgray, 
    boxrule=0.5pt, 
    colback=lightgray!20, %
    arc=3pt, 
    fontupper=\small,
    width=\textwidth, nobeforeafter, title={Instruction 3: Good, now make the robot do a moon-walk while stand up like this.},
    ]

\begin{lstlisting}[
    breaklines=true,
    escapechar=!
]
import numpy as np

reset_reward()
set_torso_targets(0.65, np.deg2rad(90), np.deg2rad(0), None, (!\colorbox{lightgreen}{-0.2}!, 0.0), 0, None)

set_feet_pos_parameters("back_left", None, None, None)
set_feet_pos_parameters("back_right", None, None, None)

!\colorbox{lightgreen}{set$\_$feet$\_$stepping$\_$parameters("back$\_$left", 0.5, 0.5, 0.5, 0.1, -0.2, True)}!
!\colorbox{lightgreen}{set$\_$feet$\_$stepping$\_$parameters("back$\_$right", 0.5, 0.5, 0.0, 0.1, -0.2, True)}!
set_feet_stepping_parameters("front_left", 0.0, 0.0, 0.0, 0.0, 0.0, False)
set_feet_stepping_parameters("front_right", 0.0, 0.0, 0.0, 0.0, 0.0, False)

execute_plan()
\end{lstlisting}

\end{tcolorbox}

\begin{tcolorbox}[
    colframe=darkgray, 
    boxrule=0.5pt, 
    colback=lightgray!20, %
    arc=3pt, 
    fontupper=\small,
    width=\textwidth, nobeforeafter, title={Instruction 4: Moon walk means the robot should walk backward while the feet swings as if they are moving forward. Correct your answer and also make it walk at a speed of 0.5m/s.},
    ]

\begin{lstlisting}[
    breaklines=true,
    escapechar=!
]
import numpy as np

reset_reward()
set_torso_targets(0.65, np.deg2rad(90), np.deg2rad(0), None, (!\colorbox{lightgreen}{-0.5}!, 0.0), 0, None)

set_feet_pos_parameters("back_left", None, None, None)
set_feet_pos_parameters("back_right", None, None, None)

set_feet_stepping_parameters("back_left", 0.5, 0.5, 0.5, 0.1, !\colorbox{lightgreen}{0.2}!, True)
set_feet_stepping_parameters("back_right", 0.5, 0.5, 0.0, 0.1, !\colorbox{lightgreen}{0.2}!, True)
set_feet_stepping_parameters("front_left", 0.0, 0.0, 0.0, 0.0, 0.0, False)
set_feet_stepping_parameters("front_right", 0.0, 0.0, 0.0, 0.0, 0.0, False)

execute_plan()
\end{lstlisting}

\end{tcolorbox}

\textbf{Dexterous Manipulator place an apple in a drawer}

\noindent\begin{tcolorbox}[
    colframe=darkgray, 
    boxrule=0.5pt, 
    colback=lightgray!20, %
    arc=3pt, 
    fontupper=\small,
    nobeforeafter, title={Instruction 1: Open the drawer.},
    ]

\begin{lstlisting}[
    breaklines=true
]
import numpy as np

reset_reward() # This is a new task so reset reward; otherwise we don't need it
set_l2_distance_reward("palm", "drawer_handle")
set_joint_fraction_reward("drawer", 1) # 1 means open

execute_plan()

\end{lstlisting}

\end{tcolorbox}

\noindent\begin{tcolorbox}[
    colframe=darkgray, 
    boxrule=0.5pt, 
    colback=lightgray!20, %
    arc=3pt, 
    fontupper=\small,
    nobeforeafter, title={Instruction 2: Good, now put the apple inside the drawer while keep it open. Assume drawer is on the floor so no need to lift up the apple.},
    ]

\begin{lstlisting}[
    breaklines=true
]
import numpy as np

reset_reward() # This is a new task so reset reward; otherwise we don't need it
set_l2_distance_reward("palm", "apple")
set_l2_distance_reward("apple", "drawer_center")
set_joint_fraction_reward("drawer", 1)

execute_plan()

\end{lstlisting}

\end{tcolorbox}

\noindent\begin{tcolorbox}[
    colframe=darkgray, 
    boxrule=0.5pt, 
    colback=lightgray!20, %
    arc=3pt, 
    fontupper=\small,
    nobeforeafter, title={Instruction 3: Good, now release the apple and move hand away.},
    ]

\begin{lstlisting}[
    breaklines=true
]
import numpy as np

reset_reward() # This is a new task so reset reward; otherwise we don't need it
set_l2_distance_reward("palm", "rest_position")

execute_plan(4)

\end{lstlisting}

\end{tcolorbox}

\noindent\begin{tcolorbox}[
    colframe=darkgray, 
    boxrule=0.5pt, 
    colback=lightgray!20, %
    arc=3pt, 
    fontupper=\small,
    nobeforeafter, title={Instruction 4: Now close the drawer.},
    ]

\begin{lstlisting}[
    breaklines=true
]
import numpy as np

reset_reward() # This is a new task so reset reward; otherwise we don't need it
set_l2_distance_reward("palm", "drawer_handle")
set_joint_fraction_reward("drawer", 0)

execute_plan()

\end{lstlisting}

\end{tcolorbox}

\end{document}